\documentclass{ieeeaccess}

\usepackage{graphicx}
\usepackage{amsmath}
\usepackage{times}
\usepackage{graphicx}
\usepackage{amssymb}
\usepackage{gensymb}
\usepackage{textcomp}
\usepackage{multirow}
\usepackage{booktabs}
\usepackage{flushend}
\usepackage{multicol}
\usepackage{color}
\usepackage{makecell}
\usepackage{bbding}
\usepackage{array}
\usepackage{pifont}
\usepackage{colortbl}
\usepackage{cite}
\usepackage{bm}
\usepackage[dvipsnames]{xcolor}

\usepackage[colorlinks,pagebackref=false,citecolor=blue,bookmarks=false,hypertexnames=true]{hyperref}
\usepackage[figureposition=bottom,tableposition=top]{caption}
\begin{document}

\history{Received XX XX 2023, accepted XX XX 2023, date of publication X 2023, date of current version XX XX 2023.}
\doi{10.1109/ACCESS.2023.X}

\title{\textit{\textbf{UnShadowNet}}: Illumination Critic Guided Contrastive Learning For Shadow Removal}
\author{\uppercase{Subhrajyoti Dasgupta\authorrefmark{1,2,\dag}, Arindam Das\authorrefmark{3,4,\dag}, Senthil~Yogamani\authorrefmark{5,\dag}, Sudip Das\authorrefmark{3}, Ciarán Eising\authorrefmark{4,6}~\IEEEmembership{Senior Member, IEEE}, Andrei Bursuc\authorrefmark{7}}, and \uppercase{Ujjwal Bhattacharya\authorrefmark{8}~\IEEEmembership{Senior Member, IEEE} }}
\address[1]{Mila - Quebec AI Institute, 6666 St-Urbain Street, Montreal, Quebec H2S 3H1, Canada.}
\address[2]{Université de Montréal, 2900 Edouard Montpetit Blvd, Montreal, Quebec H3T 1J4, Canada.}
\address[3]{Dept. Driving Software and Systems, Valeo
India, Chennai 600130, India.}
\address[4]{Dept. Electronic \& Computer Engineering, University of Limerick, Limerick, V94 T9PX, Ireland.}
\address[5]{Valeo Visions Systems, IDA Business Park, Dunmore Rd, Demesne, Galway, H54 Y276, Ireland}
\address[6]{Lero (the Science Foundation Ireland Research Centre for Software), Tierney Building, University of Limerick, Limerick, V94 NYD3, Ireland.}
\address[7]{Valeo.ai, 75017 Paris, France.}
\address[8]{Computer Vision and Pattern Recognition (CVPR) Unit, Indian Statistical Institute, Kolkata 700108, India.}
\address[$\dag$]{S. Dasgupta, A. Das, and S. Yogamani are co-first authors.}

\markboth
{S. Dasgupta \headeretal: \textit{UnShadowNet}: Illumination Critic Guided Contrastive Learning For Shadow Removal}
{S. Dasgupta \headeretal: \textit{UnShadowNet}: Illumination Critic Guided Contrastive Learning For Shadow Removal}

\corresp{Corresponding author: Ciarán Eising (e-mail: ciaran.eising@ul.ie).}

\ifx00 
\newcommand{\sy}[1]{\textcolor{red}{[SY: #1]}}
\newcommand{\AD}[1]{\textcolor{blue}{[Arindam: #1]}}
\newcommand{\andreic}[1]{\textcolor{teal}{[Andrei: \em #1]}}
\newcommand{\andrei}[1]{\textcolor{teal}{#1}}
\newcommand{\subhrajyoti}[1]{\textcolor{magenta}{#1}}
\else 
\newcommand{\sy}[1]{\textcolor{red}{}}
\newcommand{\AD}[1]{\textcolor{blue}{}}
\newcommand{\andreic}[1]{\textcolor{teal}{}}
\newcommand{\andrei}[1]{\textcolor{teal}{}}
\fi

\bstctlcite{IEEEexample:BSTcontrol}

\newcommand{\etal}{\textit{et al.}}
\newcommand{\xmark}{\ding{55}}

\begin{abstract}
Shadows are frequently encountered natural phenomena that significantly hinder the performance of computer vision perception systems in practical settings, e.g., autonomous driving. A solution to this would be to eliminate shadow regions from the images before the processing of the perception system. Yet, training such a solution requires pairs of aligned shadowed and non-shadowed images which are difficult to obtain. We introduce a novel weakly supervised shadow removal framework \textit{UnShadowNet} trained using contrastive learning.  It is composed of a \textit{DeShadower} network responsible for the removal of  the extracted shadow under the guidance of  an \textit{Illumination} network which is trained adversarially by the illumination critic and  a \textit{Refinement} network to further remove  artefacts. We show that \textit{UnShadowNet} can be easily extended to a fully-supervised set-up to exploit the ground-truth when available.  \textit{UnShadowNet} outperforms existing state-of-the-art approaches on  three publicly available shadow datasets (ISTD, adjusted ISTD, SRD) in both the weakly and fully supervised setups.

\end{abstract}

\begin{IEEEkeywords}
Shadow removal, Weakly-supervised Learning, Contrastive Learning.
\end{IEEEkeywords}

\maketitle

\section{Introduction}
Shadows are a common phenomenon that exists in most natural scenes. It occurs due to inadequate illumination that makes part of the image darker than the other region of the same image. It causes a significant negative impact on the performance of various computer vision tasks such as object detection, semantic segmentation, and object tracking. Image editing \cite{chuang2003shadow} using shadow matting is one of the common ways to remove shadows. Shadow detection and correction can improve the efficiency of the machine learning model for a broad spectrum of vision-based problems such as image restoration \cite{drew2003recovery}, satellite image analysis \cite{dare2005shadow}, information recovery in urban high-resolution panchromatic
satellite images \cite{su2016shadow}, face recognition \cite{zhang2018improving}, and object detection \cite{le2018a+}. 
In this work, we focus on natural images captured in a terrestrial setting, such as may be obtained by commercial devices and, particularly, automotive cameras.

Shadows are prevalent in almost all images in automotive scenes. The complex interaction of shadow segments with the objects of interest such as pedestrians, roads, lanes, vehicles, and riders makes the scene understanding challenging. Additionally, it does not have any distinct geometrical shape or size similar to soiling \cite{das2019soildnet, das2020tiledsoilingnet}. Thus, they commonly lead to poor performance in road segmentation \cite{dahal2021roadedgenet, chennupati2019auxnet}, pedestrian pose estimation \cite{das2022deep}, \cite{kishore2019cluenet}, \cite{das2020end}, segmentation \cite{rashedfisheyeyolo, segdasvisapp19} and trajectory prediction \cite{sharma2023navigating}. Moving shadows can be incorrectly detected as a dynamic object in background subtraction \cite{kiran2019rejection}, motion segmentation  \cite{siam2018modnet}, depth estimation \cite{kumar2021syndistnet, ravi2021fisheyedistancenet++} and SLAM algorithms \cite{gallagher2021hybrid, dahal2021online}. The difficulty of shadows is further exacerbated in strong sun glare scenes where the dynamic range is very high across shadow and glare regions \cite{yahiaoui2020let}. 
These issues lead to an incomplete or partial understanding of $360\degree$ surrounding region of the vehicle and bring major safety concerns for the passengers and Vulnerable Road Users (VRU) while performing automated driving \cite{eising2021near}.  Alternate sensor technologies like thermal camera  \cite{dasgupta2022spatio}, \cite{das2023revisiting} are resistant to shadow issues and can be used to augment cameras.


    

\begin{figure}
    \centering
    \includegraphics[width=0.488\textwidth]{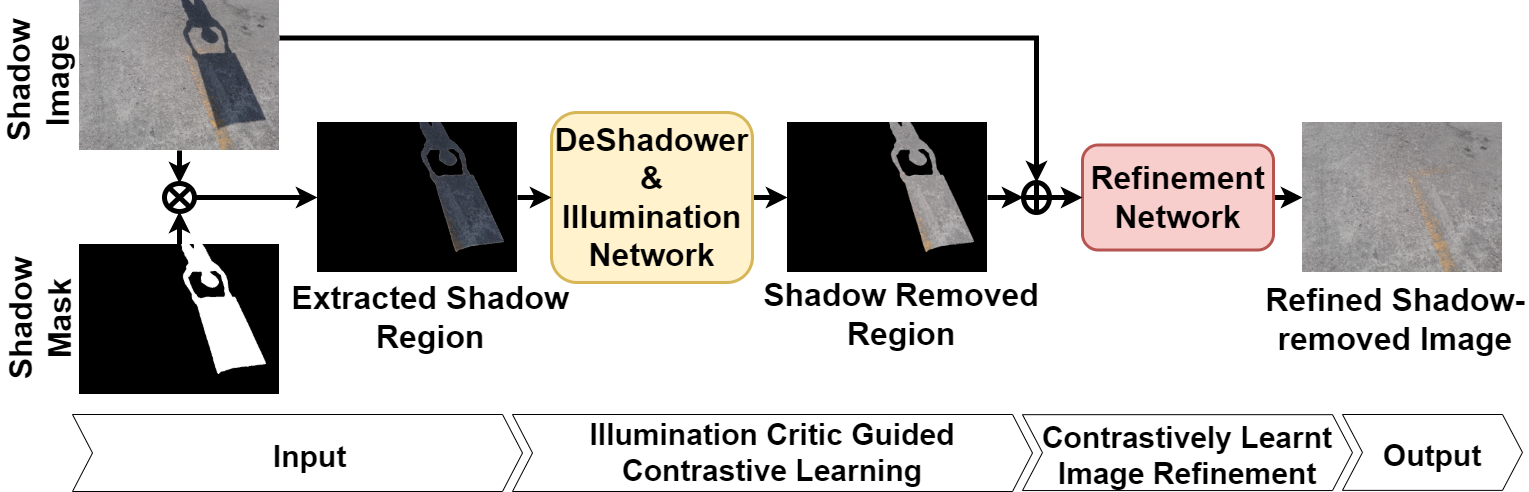}\\
    \caption{ \textbf{The proposed shadow removal framework.} The shadow image and its shadow mask are subjected to pixel-wise product operation  $\bigotimes$ to obtain the shadow extracted which is fed as input to the DeShadower ($\mathcal{D}$) and Illumination network ($\mathcal{I}$) simultaneously. $\mathcal{D}$ learns contrastively from $\mathcal{I}$ and the resultant shadow-removed region is embedded via $\bigoplus$ in the input image before feeding it to the Refinement network which produces the final Shadow-free image. The end-to-end network is trained in a weakly supervised manner.
    }


    \vspace{-7mm}
    
    \label{fig:high_level_shadow}
\end{figure}

In recent times, convolutional neural networks (CNNs) based approaches have significantly surpassed classical computer vision-based shadow removal techniques \cite{finlayson2005removal, guo2012paired, khan2015automatic, zhang2015shadow, wang2018stacked, le2019shadow}. The majority of the recent deep learning-based shadow removal approaches are fully-supervised in nature. However, such an end-to-end training setup requires \textit{paired data}, namely shadow images and their shadow-free versions of the same images. These paired data are used to train CNNs \cite{qu2017deshadownet, ding2019argan, hu2019direction}. Practically, the paired data is difficult to obtain particularly when the vehicle is moving fast. Some of the challenges include highly controlled lighting sources, object interactions, occlusions, and static scenes. Data acquisition through such a controlled setting suffers from diversity and often reports color inconsistencies \cite{wang2018stacked} between shadow and shadow-free reference of the same image. Additionally, it is very difficult to capture any High Dynamic Range (HDR) natural scene without any presence of shadow for a shadow-free reference sample.

Some of the recent studies  \cite{hu2019mask, le2020shadow, chen2021canet, liu2021shadow_lg, cun2020towards, liu2021shadow, fu2021auto, le2021physics} address the above-mentioned challenges and solve the shadow removal problem using \textit{unpaired} data. They studied the physical properties of shadows  such as illumination, color, and texture extensively. Motivated by these recent works,  we propose an end-to-end trained weakly-supervised architecture for shadow removal as illustrated in Figure \ref{fig:high_level_shadow}. 
In brief, we pass the shadow region of an input image to the DeShadower network that is aided by the Illumination network to contrastively learn to ``remove" shadow from the region by exploiting the illumination properties. It is followed by the Refinement network that helps to remove any artifacts and maintain the overall spatial consistency with the input image and finally generates a shadow-free image. 

\textbf{Summary of contributions and distinctively novel features of this work:}

\begin{enumerate}
    \item We develop a novel weakly-supervised training scheme namely \textit{UnShadowNet} using contrastive learning to build a shadow remover in unconstrained settings where the network can be trained even without any shadow-free samples.
    \item We propose a contrastive loss-guided DeShadower network to remove the shadow effects and a refinement network for efficient blending of the artifacts from shadow removed area.
    \item We achieved state-of-the-art results on three public datasets namely ISTD, adjusted ISTD, and SRD in both constrained and unconstrained setups.
    \item We perform extensive ablation studies with different proposed network components, diverse augmentation techniques, shadow inpainting, and tuning of several hyper-parameters.
\end{enumerate}

\section{Related Work}
Removing shadows from images has received a significant thrust due to the availability of large-scale datasets. In this section, first, we briefly discuss the classical computer vision methods reported in the literature. Then we discuss the more recent deep learning-based approaches. Finally, we summarize the details of contrastive learning and its applications since it is a key component in our framework.

\subsection{Classical approaches}
\textbf{Illumination-based shadow removal: }
Initial work \cite{finlayson20014, finlayson2005removal, finlayson2002removing, drew2003recovery} on removing shadows were primarily motivated by the illumination and color properties of shadow region. In one of the earliest research, Barrow \etal \cite{barrow1978recovering} proposed an image-based algorithm that decomposes the image into a few predefined intrinsic parts based on shape, texture, illumination, and shading. Later Guo \etal \cite{guo2012paired} reported the simplified version of the same intrinsic parts by establishing a relation between the shadow pixels and the shadow-free region using a linear system. Likewise, Shor \etal \cite{shor2008shadow} designed a model based on the illumination properties of shadows that makes a hard association between shadow and shadow-free pixels. In another study, Finlayson \etal \cite{finlayson2009entropy} proposed a model that generates illumination invariant image for shadow detection followed by removal. The main idea of this work is that the pixels with similar chromaticity tend to have similar albedo. Further, histogram equalization-based models performed quite well for shadow removal, where the color of the shadow-free area was transferred to the shadowed area as reported by Vicente \etal \cite{vicente2014single, vicente2017leave}.

\textbf{Shadow matting: }
Porter \& Duff \cite{porter1984compositing} introduced a matting-based technique that became effective while handling shadows that are less distinct and fuzzy around the edges. The matting method was only helpful to some extent, as computing shadow matte from a single image is difficult. To overcome this problem, Chuang \etal \cite{chuang2003shadow} applied matting for shadow editing and then transferred the shadow regions to the different scenes. Later shadow matte was computed from a sequence of video frames captured using a static camera. Shadow matte was adopted by Guo \etal \cite{guo2012paired} and Zhang \etal \cite{zhang2015shadow} in their framework for shadow removal.

\subsection{Deep learning-based approaches}
\textbf{Shadow removal using paired data: }
Deep neural networks have been able to learn the properties of a shadow region efficiently when the network is trained in a fully supervised manner. Such setup requires paired data which means the shadow and shadow-free versions of the same image are fed as input to the network. Qu \etal \cite{qu2017deshadownet} proposed an end-to-end learning framework called Deshadownet for removing shadows where they extract multi-scale contextual information from different layers. This information containing density and color offset of the shadows finally helped to predict the shadow matte. The method ST-CGAN, a two-stage approach proposed by Wang \etal \cite{wang2018stacked}, presents an end-to-end network that jointly learns to detect and remove shadows. This framework was designed based on conditional GAN \cite{isola2017image}. In SP+M-Net \cite{le2019shadow}, physics-based priors were used as inductive bias. The networks were trained to obtain the shadow parameters and matte information  to remove shadows. However, these parameters and matte details were pre-computed using the paired samples, and the same were regressed in the network. Further, Hu \etal \cite{hu2019direction} designed a shadow detection and removal technique by analyzing the contextual information in image space in a direction-aware manner. These features were then aggregated and fed into an RNN model. In ARGAN \cite{ding2019argan}, an attentive recurrent generative adversarial network was reported. The generator contained multiple steps where shadow regions were progressively detected. A negative residual-based encoder was employed to recover the shadow-free area and then a discriminator was set up to classify the final output as real or fake. In another recent framework, RIS-GAN \cite{zhang2020ris} used adversarial learning shadow removal was performed using three distinct discriminators negative residual images. Subsequently, shadow-removed images and the inverse illumination maps were jointly validated.

\textbf{Shadow removal using unpaired data: }
Mask-ShadowGAN \cite{hu2019mask} is the first deep learning-based method that learns to remove shadows from unpaired training samples. Their approach was conceptualized on CycleGAN \cite{zhu2017unpaired} where a mapping was learned from a source (shadow area) to a target (shadow-free area) domain. Le and Samaras \cite{le2020shadow} presented a learning strategy that crops the shadow area from an input image to learn the physical properties of shadow in an unpaired setting. CANet \cite{chen2021canet} handles the shadow removal problem in two stages. First, contextual information was extracted from the non-shadow area and then transferred the same to the shadow region in the feature space. Finally, an encoder-decoder setup was used to fine-tune the final results. LG-ShadowNet \cite{liu2021shadow_lg} explored the lightness and color properties of shadow images and put them through multiplicative connections in a deep neural network using unpaired data. Cun \etal \cite{cun2020towards} handled the issues of color inconsistency and artifacts at the boundaries of the shadow-removed area using a Dual Hierarchically Aggregation Network (DHAN) and a Shadow Matting Generative Adversarial Network (SMGAN). Weakly-supervised method G2R-ShadowNet \cite{liu2021shadow} designed three sub-networks dedicated to shadow generation, shadow removal, and image refinement. Fu \etal \cite{fu2021auto} modeled the shadow removal problem from a different perspective, which is auto-exposure fusion. They proposed shadow-aware FusionNet and boundary-aware RefineNet to obtain the final shadow-removed image. Further in \cite{le2021physics} a weakly-supervised approach was proposed that can be trained even without any shadow-free samples.

\textbf{Miscellaneous: }
In video sequences, cast shadows are often misinterpreted as moving objects. It was highlighted in \cite{xu2005insignificant} and considered as \textit{insignificant} shadows. These cast shadows were removed in \cite{wang2009real} by conditional random field. Liu \etal \cite{liu2011cast} investigated the cast shadows in detail by proposing a Gaussian Mixture Model at the pixel level in HSV color space followed by a pre-classifier and finally using Markov Random Fields for shadow removal. Patch-based illumination-invariant features such as binary patterns of local color constancy (BPLCC) and light-based gradient matching (LGM) were introduced in \cite{russell2017feature}. These features were used to create two dictionaries each for objects and shadows respectively. Each patch was assigned to an independent class in each iteration based on the distance from the reference dictionary. A feature fusion-based approach was followed in \cite{sahoo2021adaptive} where Spatio-Temporal Kernel Density Estimation (ST-KDE) based model was proposed for background modeling and Local Binary Pattern (LBP) features of this model were fused with the Gabor features probabilistically. Apart from shadow removal, shadow detection is also a well-studied area, some of the recent works include \cite{le2018a+, wang2020instance, chen2020multi}. Inoue \etal
\cite{inoue2020learning} highlighted the problem of preparing a large-scale shadow dataset. They proposed a pipeline to synthetically generate shadow/shadow-free/matte
image triplets.

\subsection{Contrastive learning}
Learning the underlying representations by contrasting the positive and the negative pairs have been studied earlier in the community \cite{chopra2005learning, hadsell2006dimensionality}. This line of thought has inspired several works that attempt to learn visual representations without human supervision. While one family of works uses the concept of a memory bank to store the class representations \cite{wu2018unsupervised, he2020momentum, laskin2020curl}, another set of works develops on the idea of maximization of mutual information \cite{oord2018representation, bachman2019learning, chen2020simple}. Recently, Park \etal \cite{park2020contrastive} presented an approach for unsupervised image-to-image translation by maximizing the mutual information between the two domains using contrastive learning. In our work, we adopt the problem of shadow removal to solve it without using shadow-free ground truth samples with the help of contrastive learning.

\section{Proposed Method}

\begin{figure*}
    \captionsetup{singlelinecheck=false, font=small, belowskip=-6pt}
    \centering
    \includegraphics[width=\textwidth]{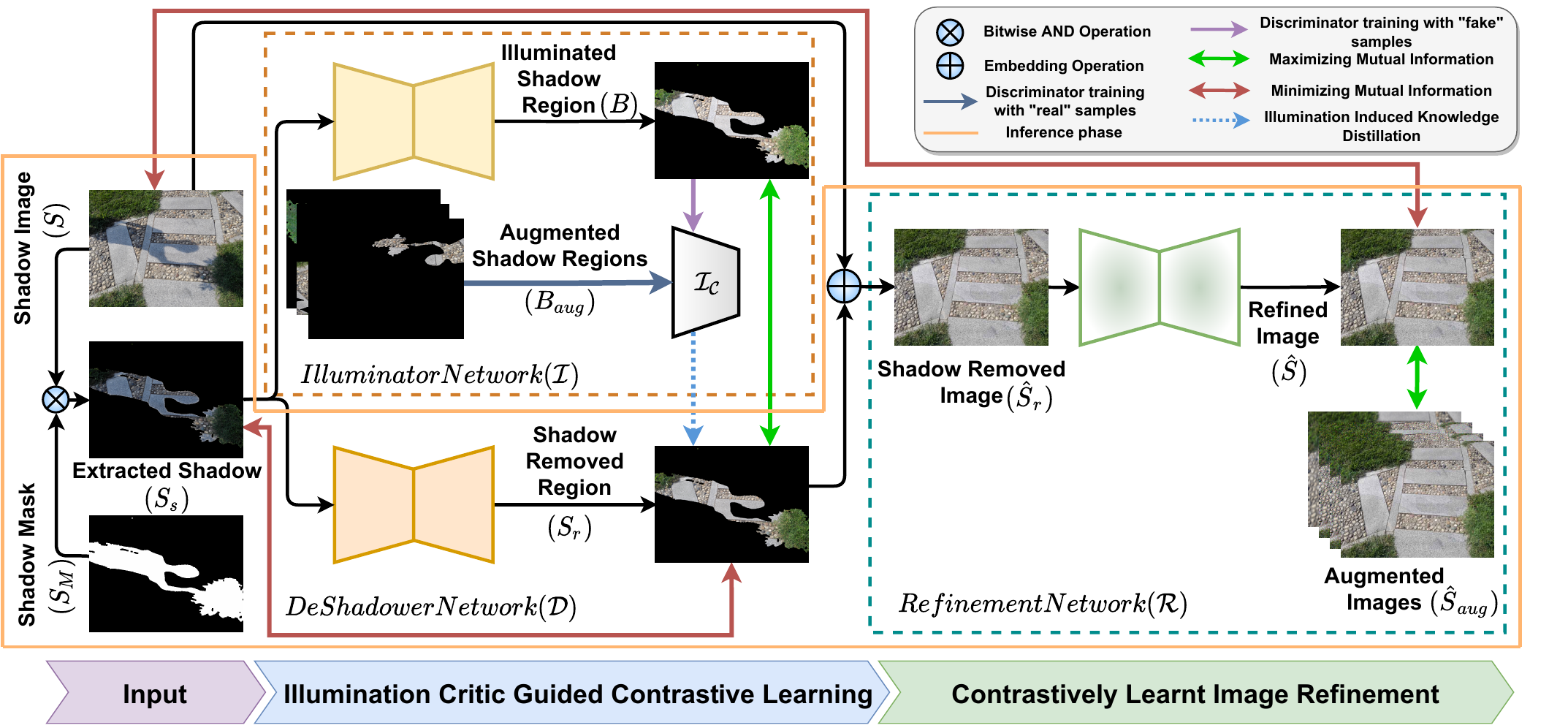}\\
    \caption{ \textit{UnShadowNet} is the proposed end-to-end weakly-supervised shadow removal architecture. It has three main sub-networks: DeShadower Network ($\mathcal{D}$), Illumination Network ($\mathcal{I}$) and Refinement Network ($\mathcal{R}$). The pixelwise product operation $\bigotimes$ between shadow image ($\mathcal{S}$) and its shadow mask ($\mathcal{S}_{M}$) extracts the shadow region ($\mathcal{S}_{s}$), which is then fed to $\mathcal{D}$ and $\mathcal{I}$ simultaneously. The generator of the adversarially trained Illumination network generates an illuminated version ($\mathcal{B}$) of $\mathcal{S}_{s}$ which is subjected to validation by a discriminator, called Illumination Critic ($\mathcal{I}_{c}$) trained on augmented shadow-free regions ($\mathcal{B}_{aug}$). DeShadower is trained to produce shadow-removed region ($\mathcal{S}_{r}$) of $\mathcal{S}_{s}$. To create a more realistic illumination region $\mathcal{S}_{r}$, a contrastive approach is employed between $\mathcal{B}$ and $\mathcal{S}_{r}$. Finally, shadow-removed image ($\hat{S}_r$) is obtained by applying embedding operation $\bigoplus$ to become input to the Refinement network. $\mathcal{R}$ is trained to efficiently blend the areas between shadow-removed and non-shadow regions so that it is robust to noise, blur, etc. Here contrastive learning approach was followed where \textit{positive} samples ($\hat{S}_{aug}$) were generated as per the method in\cite{chen2020simple}.}
    
    \label{fig:self_sup_shadow_arch}
\end{figure*}

\definecolor{LightCyan}{rgb}{0.88,1,1}
\definecolor{LightGreen}{rgb}{0.35,0.8,0}
\definecolor{LightGreen2}{rgb}{0.5,0.9,0}
\definecolor{Gray}{gray}{0.85}



In this work, we define the problem of shadow removal as the translation of images from the shadow domain $\mathcal{S} \subset {\mathbb{R}^{H \times W \times C}}$ 
to shadow-free domain $\mathcal{F} \subset {\mathbb{R}^{H \times W \times C}}$ by utilizing only the shadow image and its mask and alleviating the use of its shadow-free counterpart. The proposed architecture \textit{UnShadowNet} is illustrated in Figure \ref{fig:self_sup_shadow_arch}. We briefly summarize the high-level characteristics here and discuss each part in more detail in the following subsections. In this section, we present the overall architecture of our proposed end-to-end shadow removal network, namely   \textit{UnShadowNet}. The architecture can be divided into $three$ parts: DeShadower Network ($\mathcal{D}$), Illumination Network ($\mathcal{I}$) and Refinement Network ($\mathcal{R}$). These three networks are jointly trained in a weakly-supervised manner. Let us consider a shadow image $S\in\mathcal{S}$ and its corresponding shadow mask $S_M$. We obtain the shadow region $S_s$ by cropping the masked area from $S_M$ in the shadow image $S$. The DeShadower Network learns to remove the shadow from the region using a contrastive learning setup. It is aided by the Illumination Network which generates bright samples for $\mathcal{D}$ to learn from. The Refinement Network finally combines the shadow-free region $S_f$ with the real image and refines it to form the shadow-free image $\hat{S}$.

\subsection{DeShadower Network ($\mathcal{D}$)}
The DeShadower Network is designed as an encoder-decoder-based architecture
that generates a shadow-removed region ($S_r$) from the shadow region ($S_s$). The shadow-removed regions generated by this network $S_r$ should \textit{associate} more with the bright samples and 
\textit{dissociate} itself from the shadow samples. We employ a contrastive learning approach to help the DeShadower network achieve this and learn to generate shadow-free regions.  In a contrastive learning framework, a \textit{``query"} maximizes the mutual information with a \textit{``positive"} sample in contrast to other samples that are referred to as \textit{``negatives"}.
In this work, we use a ``noise contrastive estimation" framework \cite{oord2018representation} to maximize the mutual information between $S_f$ and the bright sample $B$. We treat the bright samples generated by the Illumination Network as the ``positive" and the shadow regions as the ``negatives" in this contrastive learning setup. Thus, the objective function for maximizing (and minimizing) the mutual information can be formulated with the InfoNCELoss~\cite{oord2018representation}, a criterion derived from both statistics~\cite{gutmann2010noise, oord2018representation} and metric learning~\cite{chopra2005learning, hadsell2006dimensionality, sohn2016improved}. Its formulation bears similarities with the cross-entropy loss:
\begin{multline}
    \ell(x, x^+, x^-) = \\ - log \left[ \frac{exp(x\cdot x^+/ \tau)}{exp(x\cdot x^+ / \tau) + \sum_{i=1}^N exp(x\cdot x^-_i/ \tau)} \right]
\end{multline}
where $x, x^+, x^-$ are the \textit{query}, \textit{positive} and \textit{negatives} respectively. 
$\tau$ is the temperature parameter that controls the sharpness of the similarity distribution. We set it to the default value from prior work~\cite{he2020momentum, wu2018unsupervised}: $\tau {=} 0.07$.

The feature stack in the encoder of the DeShadower Network, represented as $\mathcal{D}_{enc}$, already contains latent information about the input shadow region $S_s$. From $\mathcal{D}_{enc}$, $L$ layers are selected, and following practices from prior works~\cite{chen2020simple}, we pass these features through a projection head, an MLP ($M_l$) with two hidden layers.
Subsequently, we obtain features: 
\begin{equation}
    s_l = M_l(\mathcal{D}^l_{enc}(S_s)); \quad l \in \{1,2,...,L\}
\end{equation}
where $\mathcal{D}^l_{enc}$ is the $l$-th chosen layer in $\mathcal{D}^l_{enc}$. Similarly the output or the `unshadowed' region $S_f$ and the bright region $B$ are encoded respectively as:
\begin{equation}
    f_l = M_l(\mathcal{D}^l_{enc}(S_f)); b_l = M_l(\mathcal{D}^l_{enc}(B))
\end{equation}  
We 
adjust the InfoNCE loss \cite{oord2018representation} into a layer-wise NCE loss:
\begin{equation}
    \mathcal{L}_{NCE}(f_l, b_l, s_l) = \mathbb{E}_{S_f \sim \mathcal{F}, S_s \sim \mathcal{S}, B \sim \mathcal{B}} \quad \ell(f_l, b_l, s_l)
\end{equation}

The generator should not change the contents of an image when there is no need to. In other words, given a shadow-free sample as input, it is expected to generate the same output without any change. To enforce such a regularization, we employ an identity loss \cite{zhu2017unpaired, taigman2016unsupervised}. 
It is formulated using an $L1$ loss as:
\begin{equation}
    \mathcal{L}_{iden} = \mathbb{E}_{S_f \sim \mathcal{F}} ||\mathcal{D}(S_f), S_f||_1
\end{equation}

Additionally, as described further in the following sections, the Illumination Critic $\mathcal{I}_\mathcal{C}$ is trained on real non-shadow samples and augmented bright samples. Therefore, we can additionally use the cues provided by the Illumination Critic to distil its knowledge of illumination to the DeShadower Network. This is achieved by computing the loss:
\begin{equation}
\label{eqn:critic}
\mathcal{L}_{\textit{critic}} = 
\left[ 1 - \mathcal{I}_\mathcal{C}(\mathcal{D}(S_s)) \right]^2 
\end{equation}

\subsection{Illumination Network ($\mathcal{I}$)}
\begin{figure}
    \centering
    
    \includegraphics[width=0.11\textwidth]{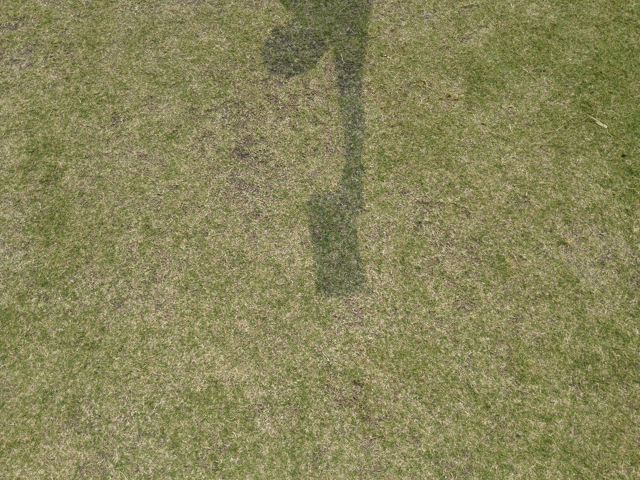}
    \includegraphics[width=0.11\textwidth]{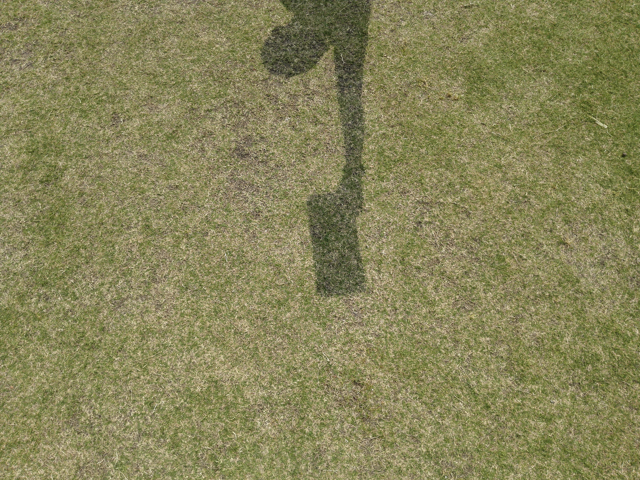}
    \includegraphics[width=0.11\textwidth]{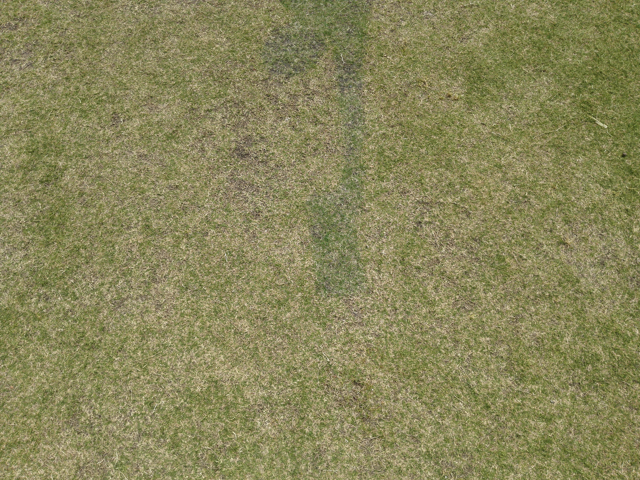}
    \includegraphics[width=0.11\textwidth]{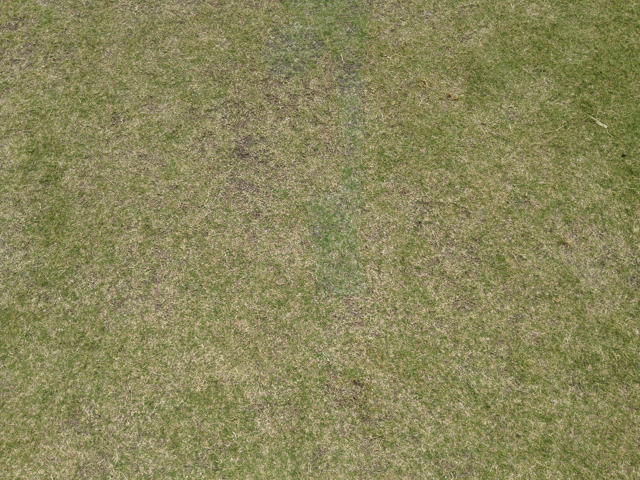}
   \hspace{3cm}{Original} \hspace{1cm}{Shadow with varying illumination level}
    \caption{
    Illustration of different illumination control levels of shadow region. 
    }
 \vspace{-9 mm}
 
    \label{fig:varying_shadow}
\end{figure}

Shadow regions have a lower level of illumination compared to their surroundings. The exact illumination level can vary according to scene lighting conditions  as illustrated in Fig. \ref{fig:varying_shadow}. To show that a real shadow image and an image with a region where brightness is reduced are similar even semantically, we designed a small experimental setup. We fine-tune a ResNet \cite{he2016deep} with samples containing real shadows and no shadows for a Shadow/Non-shadow classification task and then test the images where we reduce the brightness in the shadow region. In the majority of the cases, the network classifies it to be a `Shadow' image. 

Using this heuristic, the Illumination Network ($\mathcal{I}$) is designed as a Generative Adversarial Network \cite{goodfellow2014generative} to serve as a complementary augmentation setup to generate synthetic images where the illumination level is increased in a shadow region. The shadow region $S_s$ is passed through the generator $\mathcal{I}_\mathcal{G}$ to produce brighter samples $B$ of the shadow region.  The illumination critic ($\mathcal{I}_\mathcal{C}$) learns to classify these samples generated by $\mathcal{I}_\mathcal{G}$ as `fake'. The motivation of this discriminator is detailed in the following section. The generator $\mathcal{I}_\mathcal{G}$ and the discriminator $\mathcal{I}_\mathcal{C}$ thus learns from the adversarial loss as:

\begin{equation}
\label{eqn:adversarial}
\begin{aligned}
    \mathcal{L}_{\textit{adv}} = {} & \mathbb{E}_{S_s \sim \mathcal{S}} \left[ (1 - \mathcal{I}_\mathcal{C}(\mathcal{I}_\mathcal{G}(S_s)))^2 \right] + \\
        & \mathbb{E}_{S_s \sim \mathcal{S}} \left[ \mathcal{I}_\mathcal{C}(\mathcal{I}_\mathcal{G}(S_s))^2 \right] + \\ 
        & \mathbb{E}_{B' \sim \mathcal{B}} \left[ (1 - \mathcal{I}_\mathcal{D}(B'))^2 \right]
\end{aligned}
\end{equation}

We observe that the more optimal samples the Illumination Network generates, the better it aids $\mathcal{D}$ to create more realistic shadow-removed samples. Therefore, to improve $\mathcal{I}$ to create well-illuminated samples we employ the illumination loss as an $L1$ loss between the $\mathcal{I}_\mathcal{G}$ generated bright sample $B$ and the shadow-removed sample $S_f$ as:
\begin{equation}
    \mathcal{L}_{illum} = \frac{1}{N}\sum_{i=0}^{N}||S_f - B||_1
\end{equation}
The adversarial loss with the help of the discriminator and the illumination loss together play a role in generating well-illuminated samples, which in turn helps $\mathcal{D}$ to create better shadow-removed samples. In this regard, both $\mathcal{D}$ and $\mathcal{I}$ complement each other for the task. The Illumination Network supervises $\mathcal{D}$  to generate shadow-removed regions and likewise, $\mathcal{D}$ encourages $\mathcal{I}$ to create well-illuminated samples by learning from it. 
The choice of using $\mathcal{I}$ is experimentally justified in the ablation study section, as it helps to generate better results rather than relying solely on a pre-determined illumination level increase.

\subsection{Illumination Critic ($\mathcal{I}_\mathcal{C}$)}
\label{critic}

The role of the Illumination Critic ($\mathcal{I}_\mathcal{C}$)  is two-fold. 
Firstly, in the Illumination Network which  generates well-illuminated variations of the shadow region $S_s$, the $\mathcal{I}_\mathcal{C}$ is designed as a discriminator to the $\mathcal{I}_\mathcal{G}$. The knowledge $\mathcal{I}_\mathcal{C}$ learns from representations of shadow-free regions allows it to encourage $\mathcal{I}_\mathcal{G}$ to create well-illuminated variations of the shadow region $S_s$ which is later used as \textit{positive} pair to contrastively train  $\mathcal{D}$.

Additionally, the DeShadower Network utilizes the knowledge of the $\mathcal{I}_\mathcal{C}$ to create realistic shadow-removed regions from the $S_s$. Having learned the representations of shadow-free regions and augmented samples with varying illumination, $\mathcal{I}_\mathcal{C}$ can influence $\mathcal{D}$ to ``remove" shadows from shadow regions using the $\mathcal{L}_{critic}$ in Eqn. \ref{eqn:critic}. This two-fold characteristic of $\mathcal{I}_\mathcal{C}$ facilitates the complementary nature of $\mathcal{D}$ and $\mathcal{I}$ where they mutually improve each other. 

To train $\mathcal{I}_\mathcal{C}$, we crop randomly masked non-shadow areas from $S$ as well as other samples in the dataset similar to \cite{liu2021shadow}. Additionally, $\mathcal{I}_\mathcal{C}$ is  trained by augmented samples where each shadow region $S_s$ is converted to $3$ different samples by varying the illumination levels. The illumination levels are increased by a factor $\mu-5, \mu, \mu+5$ where $\mu$ is fixed empirically as presented in Table \ref{tab:mu_table}. It is trained using the same adversarial loss as the Illumination Network.

\vspace{-3mm}

\subsection{Refinement Network ($\mathcal{R}$)} 
After obtaining the shadow-removed region $S_r$, it is embedded with the original shadow image $S$.  
The embedding operation can be defined as:
\begin{equation}
    \hat{S}_r = S - S * S_M + S_r * S_M
\end{equation}
Following the embedding operation, there remain additional artefacts around the inpainted area due to improper blending. The Refinement Network $\mathcal{R}$ is designed to get rid of such artefacts by making use of the global context in the image. The absence of explicit ground truths in this setting motivated us to design a contrastive setup to train $\mathcal{R}$. 
To generate the \textit{positive} samples, we follow \cite{chen2020simple} to augment the generated shadow-removed image ($\hat{S}_r$) by using random cropping of non-shadow regions. It is followed by additional transformations like resizing the cropped region back to the original size, random cutout, Gaussian blur, and Gaussian noise, represented as $\hat{S}_{aug}$. The objective is to maximize the information between the \textit{query} image and the \textit{positive} image pairs and reduce the same with the \textit{negative} ones. In this phase, we reuse the existing encoder of $\mathcal{R}$ represented as $\mathcal{R}_{enc}$ as a feature extractor. We extract the layer-wise features of the \textit{query} ${F}_{l}$, \textit{positive} $F^+_l$ and \textit{negative} $F^-_l$ images and pass them through an MLP with two-hidden layers, similar to $\mathcal{D}$. Thus, we obtain the feature representations of ${F}_{l}$, $F^+_l$ and $F^-_l$ respectively as follows: 
\begin{equation}
\begin{split}
{F}_{l} = \hat{M}_l(\mathcal{R}^l_{enc}(\hat{S}));   
F^+_l = \hat{M}_l(\mathcal{R}^l_{enc}(\hat{S}_{aug})); \\
F^-_l = \hat{M}_l(\mathcal{R}^l_{enc}(S))
\end{split}
\end{equation}

Therefore the objective function for the contrastive learning setup can be represented as:
\begin{equation}
    \mathcal{L}_{NCE}(F_l, F^+_l, F^-_l) = \mathbb{E}_{\hat{S} \sim \mathcal{F}, S \sim \mathcal{S}} \quad \ell(F_l, F^+_l, F^-_l)
\end{equation}

Additionally, we find that following \cite{xu2021learning, johnson2016perceptual}, using a ``layer-selective" perceptual loss along with the contrastive loss helps to preserve the integrity of the overall spatial details present in the input and output images. It is computed based on the features extracted by {\fontfamily{qcr}\selectfont
relu\_5\_1
}and
{\fontfamily{qcr}\selectfont
relu\_5\_3
}of a
VGG-16 \cite{simonyan2014very} feature extractor as:

\begin{equation}
    \mathcal{L}_{ref} = \frac{1}{2}\sum_{i=0}^{2}||VGG_i(\hat{S}) - VGG_i(S)||^2_2
\end{equation}

\subsection{Supervised setup}
Paired data is difficult to obtain for large-scale real-world datasets, however, it can be collected for a controlled smaller dataset. Here we demonstrate that \textit{UnShadowNet} can be easily extended to exploit when paired shadow-free ground-truths ($G$) are available. Since the optimal level of illumination in the regions are available from $G$ itself, we remove $\mathcal{I}$ in the fully-supervised setup and use different augmented versions of the $G$ directly. Additionally, we make use of different losses that help to generate more realistic shadow-free images. To avoid loss of details in terms of content \cite{ledig2017photo}, we employ the pixel-wise \textit{L1}-norm:  
\begin{equation}
    \mathcal{L}_{p} = \frac{1}{N} \sum_{i=0}^{N} ||\hat{S}_i - G_i||
\end{equation}

\begin{table*}[!h]
\centering

\scalebox{1}{
\begin{tabular}{l|c|c|c|c|c|c|c|c|c} 
\bottomrule

  &  \multicolumn{3}{c|} {\textbf{Shadow Region}} & \multicolumn{3}{c|}{\textbf{Non-Shadow Region}} & \multicolumn{3}{c}{\textbf{All}}\\
\cline{1-10}
\cellcolor{blue!25} $\textbf{Methods}$  &  \cellcolor{red!25}$\textbf{RMSE}$$\downarrow$  & \cellcolor{red!25}$\textbf{PSNR}$$\uparrow$ & \cellcolor{red!25}$\textbf{SSIM}$$\uparrow$ & \cellcolor{red!25}$\textbf{RMSE}$$\downarrow$  & \cellcolor{red!25}$\textbf{PSNR}$$\uparrow$ & \cellcolor{red!25}$\textbf{SSIM}$$\uparrow$ & \cellcolor{red!25}$\textbf{RMSE}$$\downarrow$  & \cellcolor{red!25}$\textbf{PSNR}$$\uparrow$ & \cellcolor{red!25}$\textbf{SSIM}$$\uparrow$  \\
\cline{1-10}

\midrule

\rowcolor{LightCyan}
\multicolumn{10}{c}{\textbf{Weakly-Supervised Framework}}
\\
\midrule

D-Net  & 11.8 & 32.46 & 0.978 & 4.6 & 34.85 & 0.972 & 6.1 & 30.26 & 0.947\\

D+I-Net  & 9.2 & 33.68 & 0.981  & 3.2 & 35.03 & 0.974  & 4.7 & 30.28 & 0.949 \\ 

D+R-Net  & 9.9 & 33.43 & 0.980  & 3.4 & 35.47 & 0.974  & 5.0 & 30.17 & 0.950 \\ 

D+I+R-Net  & 8.9 & 34.01 & 0.982  & 2.9 & 35.48 & 0.976  & 4.4 & 30.41 & 0.951 \\ 

\textbf{UnShadowNet}  & \textbf{8.3} & \textbf{34.47} & \textbf{0.983} & \textbf{2.9} & \textbf{35.51} & \textbf{0.977} & \textbf{3.8} & \textbf{30.63} & \textbf{0.951} \\ 

\midrule
\midrule
\rowcolor{LightCyan}
\multicolumn{10}{c}{\textbf{Fully-Supervised Framework}}
\\
\midrule

D-Net  & 7.7 & 35.57 & 0.984 & 4.6 & 35.24 & 0.972 & 5.2 & 31.24 & 0.952\\


D+R-Net  & 6.3 & 36.13 & 0.989  & 2.8 & 36.03 & 0.978  & 3.8 & 31.76 & 0.958 \\ 


\textbf{UnShadowNet \textit{Sup.}} & \textbf{5.9}  & \textbf{36.19} & \textbf{0.989} & \textbf{2.7}  & \textbf{36.44} & \textbf{0.978} & \textbf{3.3}  & \textbf{31.98} & \textbf{0.959}  \\ 
\cline{1-10}

\toprule

\end{tabular}
}

\caption{\textbf{Ablation study of the various components of UnShadowNet} in both weakly-supervised and fully-supervised setup on ISTD \cite{wang2018stacked} dataset using RMSE, PSNR and SSIM metrics.}
\label{tab:shadow_network_ablation}
\vspace{-5mm}

\end{table*}

Color plays an important role in preserving the realism of the generated image and maintaining consistency with the real image. To this end, we follow a recent study in the literature \cite{wang2019underexposed} to formulate the color loss as:

\begin{equation}
    \mathcal{L}_{c} = \frac{1}{N} \sum_{i=0}^{N}\sum_{j=0}^{P} \angle (\hat{S}_i, G_i)
\end{equation}
where $\angle(,)$ computes an angle between two colors regarding the RGB color as a $3$D vector \cite{wang2019underexposed}, and $P$ represents the number of pixel-pairs.

In addition, style plays an important role in an image that corresponds to the texture information \cite{gatys2015texture}. We follow \cite{gatys2016image} to define a Gram matrix as the inner product between
the vectorised feature maps $i$ and $j$ in layer $l$: 
\begin{equation}
    \gamma_{i,j}^l = \sum_k V_{i,k}^l \cdot V_{j,k}^l
\end{equation}
The Gram matrix is the style for the feature set extracted by the $l$-th layer of VGG-$16$ net for an input image. Subsequently, the style loss can be defined as:
\begin{equation}
    \mathcal{L}_{s} = \frac{1}{N_l} \sum_{i = 0}^{N_l} ||\hat{S}_i - \gamma_i||^2
\end{equation}
where $S_i$ and $\gamma_i$ are the gram matrices for the generated shadow-free image and ground truth image respectively using VGG-$16$.

Therefore, the complete supervised loss can be formulated as a weighted sum ($\mathcal{L}_{sup}$) of the pixel ($\mathcal{L}_{p}$), color ($\mathcal{L}_{c}$) and style ($\mathcal{L}_{s}$) losses:
\begin{equation}
    \mathcal{L}_{sup} = \lambda_1 \cdot \mathcal{L}_{p} + \lambda_2 \cdot \mathcal{L}_{c} + \lambda_3 \cdot \mathcal{L}_{s}
\end{equation}

where $\lambda_1,\lambda_2$ and $\lambda_3$ are the weights corresponding to the pixel, color, and style losses respectively and are set empirically to $1.0$, $1.0$ and $1.0 \times 10^{4}$ following \cite{li2018single}, \cite{wang2019underexposed} and \cite{gatys2016image} respectively in our experiments.

\section{Experimentation Details}  \label{sec:performance}

\definecolor{LightCyan}{rgb}{0.88,1,1}
\definecolor{LightGreen}{rgb}{0.35,0.8,0}
\definecolor{LightGreen2}{rgb}{0.5,0.9,0}
\definecolor{Gray}{gray}{0.85}

\begin{table*}[!h]

\centering

\scalebox{0.72}{
\begin{tabular}{c|c|c|c|c|c|c|c|c|c|c|c} 
\bottomrule
  & & &  \multicolumn{3}{c|} {\textbf{Shadow Region}} & \multicolumn{3}{c|}{\textbf{Non-Shadow Region}} & \multicolumn{3}{c}{\textbf{All}}\\


\cellcolor{blue!25}\textbf{Curriculum Learning} & 
\cellcolor{blue!25}\textbf{Shadow Inpainting} &
\cellcolor{blue!25}\textbf{Data Augmentation} &
\cellcolor{red!25}\textbf{RMSE} $\downarrow$  & 
\cellcolor{red!25}\textbf{PSNR} $\uparrow$ & 
\cellcolor{red!25}\textbf{SSIM} $\uparrow$ & 
\cellcolor{red!25}\textbf{RMSE} $\downarrow$  & 
\cellcolor{red!25}\textbf{PSNR} $\uparrow$ & 
\cellcolor{red!25}\textbf{SSIM} $\uparrow$ & 
\cellcolor{red!25}\textbf{RMSE} $\downarrow$  & 
\cellcolor{red!25}\textbf{PSNR} $\uparrow$ & 
\cellcolor{red!25}\textbf{SSIM} $\uparrow$  \\

\midrule
\rowcolor{LightCyan}
\multicolumn{12}{c}{\textbf{Weakly-Supervised Framework}}
\\
\midrule

\color{green}\Checkmark & \color{red}\xmark & \color{red}\xmark & 9.08 & 33.88 & 0.983 & 3.65 & 35.34 & 0.977 & 3.99 & 30.08 & 0.949\\ 

\color{green}\Checkmark & \color{green}\Checkmark & \color{red}\xmark & 8.65 & 34.12 & 0.984 & 3.11 & 35.47 & 0.977 & 3.87 & 30.41 & 0.951\\  

\color{green}\Checkmark & \color{red}\xmark & \color{green}\Checkmark & 9.01 & 33.91 & 0.983 & 3.54 & 35.36 & 0.977 & 3.96 & 30.15 & 0.950\\ 

\color{red}\xmark & \color{green}\Checkmark & \color{red}\xmark & 8.97 & 34.01 & 0.983 & 3.42 & 35.40 & 0.977 & 3.96 & 30.23 & 0.950\\ 

\color{red}\xmark & \color{red}\xmark & \color{green}\Checkmark & 9.23 & 33.59 & 0.981 & 3.74 & 35.22 & 0.976 & 4.08 & 29.88 & 0.947\\

\color{red}\xmark & \color{green}\Checkmark & \color{green}\Checkmark & 8.89 & 34.03 & 0.983 & 3.38 & 35.43 & 0.977 & 3.94 & 30.32 & 0.950\\

\color{green}\Checkmark & \color{green}\Checkmark & \color{green}\Checkmark & \textbf{8.31} & \textbf{34.47} & \textbf{0.984} & \textbf{2.92} & \textbf{35.51} & \textbf{0.977} & \textbf{3.80} & \textbf{30.63} & \textbf{0.951}\\

\midrule
\midrule
\rowcolor{LightCyan}
\multicolumn{12}{c}{\textbf{Fully-Supervised Framework}}
\\
\midrule

\color{green}\Checkmark & \color{red}\xmark & \color{red}\xmark & 6.90 & 35.75 & 0.986 & 3.03 & 36.20 & 0.977 & 3.76 & 31.08 & 0.958\\ 

\color{green}\Checkmark & \color{green}\Checkmark & \color{red}\xmark & 6.23 & 36.12 & 0.989 & 2.82 & 36.38 & 0.978 & 3.58 & 31.17 & 0.959\\  

\color{green}\Checkmark & \color{red}\xmark & \color{green}\Checkmark & 6.86 & 35.92 & 0.986 & 2.99 & 36.12 & 0.977 & 3.74 & 31.01 & 0.958\\ 

\color{red}\xmark & \color{green}\Checkmark & \color{red}\xmark & 6.81 & 36.07 & 0.987 & 3.01 & 36.21 & 0.977 & 3.67 & 31.27 & 0.959\\ 

\color{red}\xmark & \color{red}\xmark & \color{green}\Checkmark & 7.13 & 34.82 & 0.982 & 3.28 & 36.02 & 0.976 & 3.93 & 31.46 & 0.956\\

\color{red}\xmark & \color{green}\Checkmark & \color{green}\Checkmark & 6.47 & 36.09 & 0.987 & 2.96 & 36.26 & 0.978 & 3.63 & 31.85 & 0.959\\

\color{green}\Checkmark & \color{green}\Checkmark & \color{green}\Checkmark & \textbf{5.92} & \textbf{36.20} & \textbf{0.989} & \textbf{2.71} & \textbf{36.44} & \textbf{0.978} & \textbf{3.33} & \textbf{31.98} & \textbf{0.959}\\
\bottomrule
\end{tabular}
}
\caption{\textbf{Ablation study of UnShadowNet using different training strategies on adjusted ISTD dataset.} 
}
\label{tab:ablation_data_aug}
\vspace{-7mm}
\end{table*}

\subsection{Dataset and evaluation metrics}
\textbf{Datasets:} In this work, we train and evaluate our proposed method on three publicly available datasets discussed below.

\textbf{ISTD:} ISTD \cite{wang2018stacked} contains image triplets: a shadow image, a shadow mask, and a shadow-free image captured at different lighting conditions that make the dataset significantly diverse. A total of $1,870$ image triplets were generated from $135$ scenes for the training set, whereas the testing set contains $540$ triplets obtained from $45$ scenes.

\textbf{ISTD+:} The samples of ISTD \cite{wang2018stacked} dataset were found to have color inconsistency issues between the shadow and shadow-free images as mentioned in the original work \cite{wang2018stacked}. The reason was that shadow and shadow-free image pairs were collected at different times of the day which led to the effect of different lighting appearance in the images. This color irregularity issue was fixed by Le \etal \cite{le2019shadow} and an adjusted ISTD (ISTD+) dataset was published. 

\textbf{SRD:} There are total $408$ pairs of shadow and shadow-free images in SRD \cite{qu2017deshadownet} dataset without the shadow-mask. For the training and evaluation of our both constrained and unconstrained setup, we use the shadow masks publicly provided by Cun \etal \cite{cun2020towards}.\\

\textbf{Evaluation metrics:} For all the experiments conducted in this work, we use Root Mean-Square Error (RMSE), Peak Signal-to-Noise Ratio (PSNR), and Structural Similarity (SSIM) respectively as metrics to evaluate and compare the proposed approach with other state-of-the-art methods. Following the prior-art \cite{guo2012paired, qu2017deshadownet, wang2018stacked, le2019shadow, hu2019mask, le2020shadow, liu2021shadow_lg}, we compute the RMSE on the recovered shadow-free area, non-shadow area and the entire image in LAB color space.
In addition to RMSE, we also compute PSNR and SSIM scores in RGB color space. RMSE is interpreted as better when it is lower, while PSNR and SSIM are better when they are higher.

\begin{figure*}[]

    \centering
    
    \includegraphics[width=0.18\textwidth]{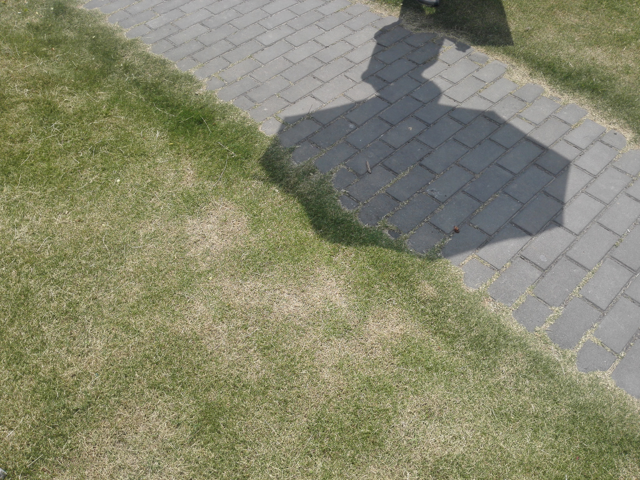}
    \includegraphics[width=0.18\textwidth]{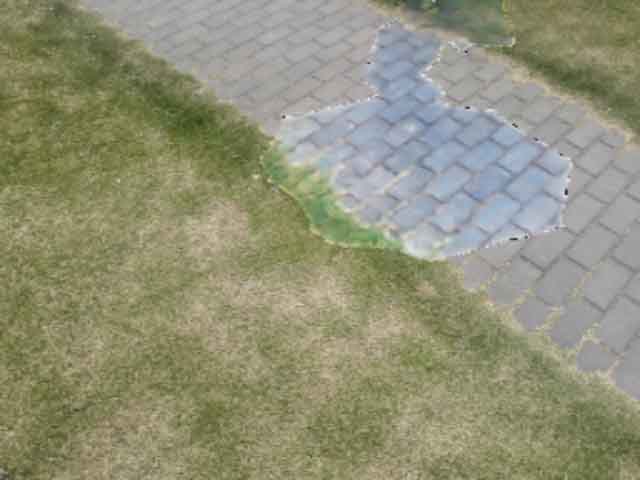}
    \includegraphics[width=0.18\textwidth]{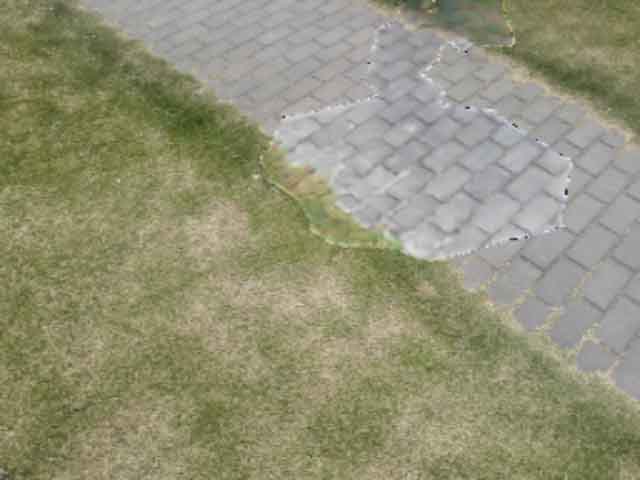}
    \includegraphics[width=0.18\textwidth]{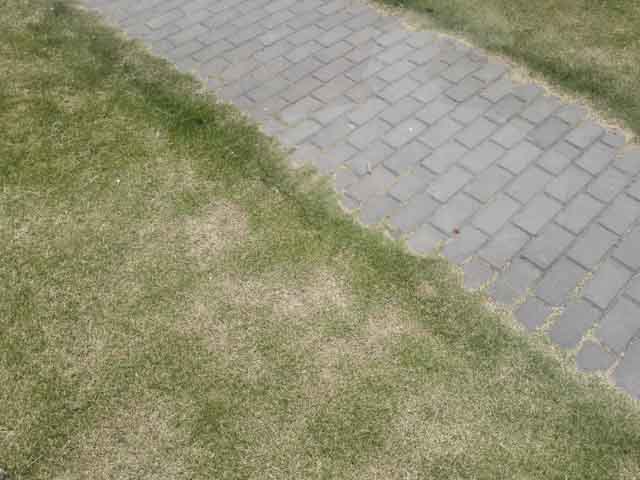}
    \includegraphics[width=0.18\textwidth]{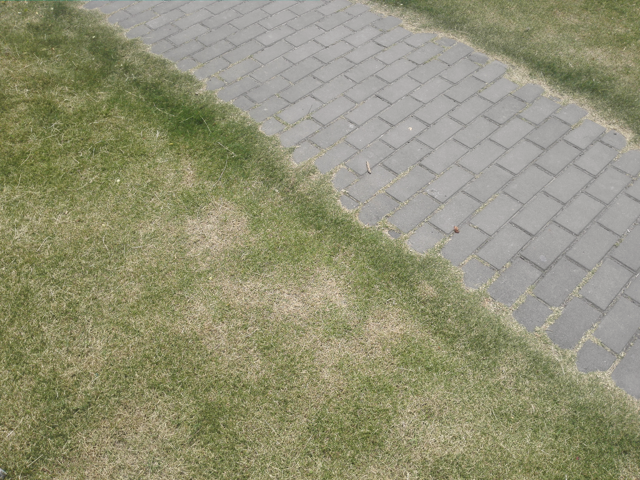} \\
    \vspace{0.4 mm}
    \includegraphics[width=0.18\textwidth]{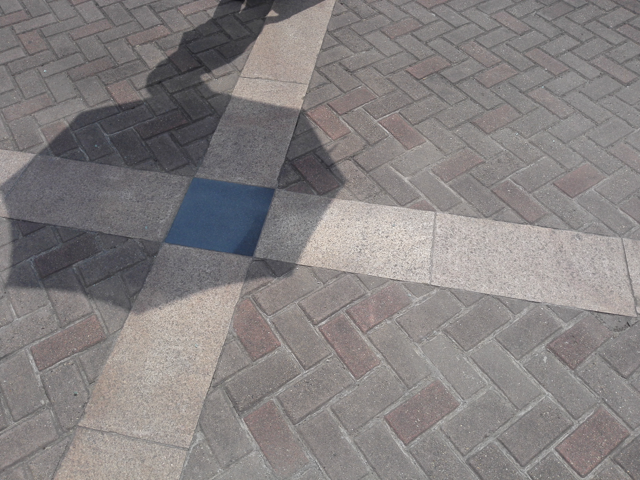}
    \includegraphics[width=0.18\textwidth]{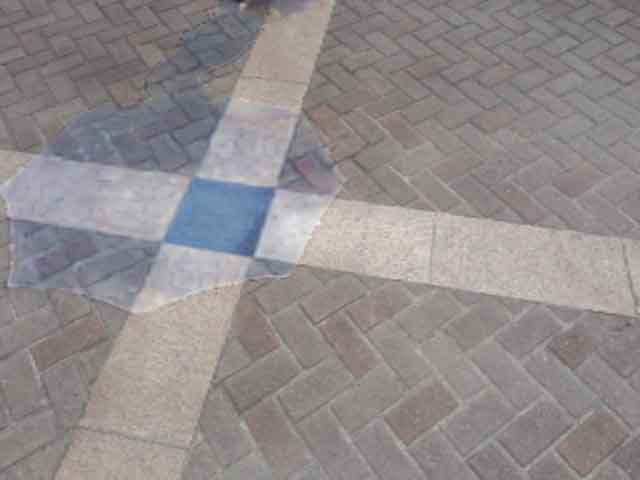}
    \includegraphics[width=0.18\textwidth]{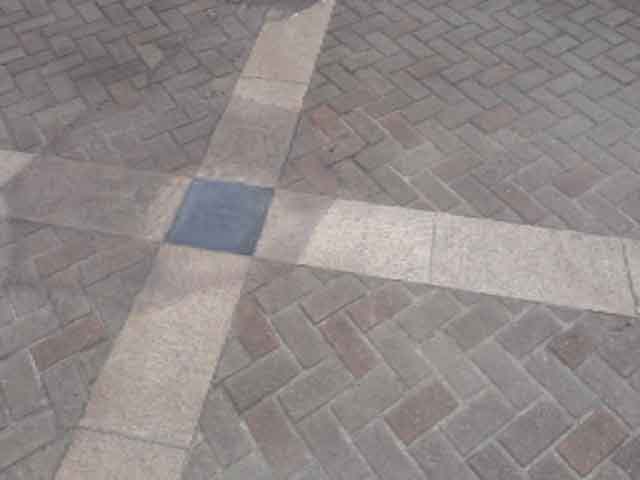}
    \includegraphics[width=0.18\textwidth]{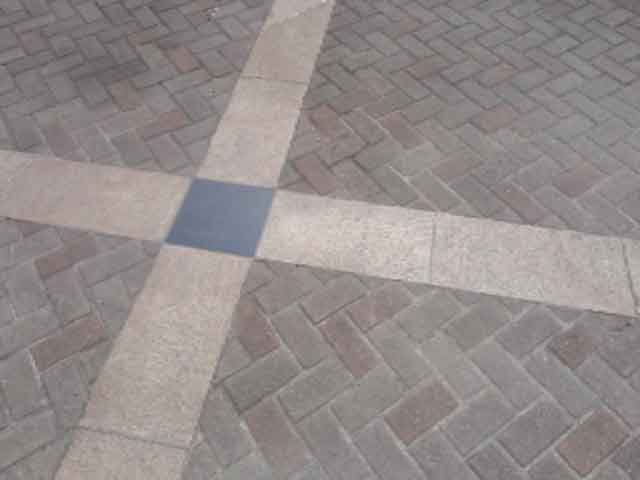}
    \includegraphics[width=0.18\textwidth]{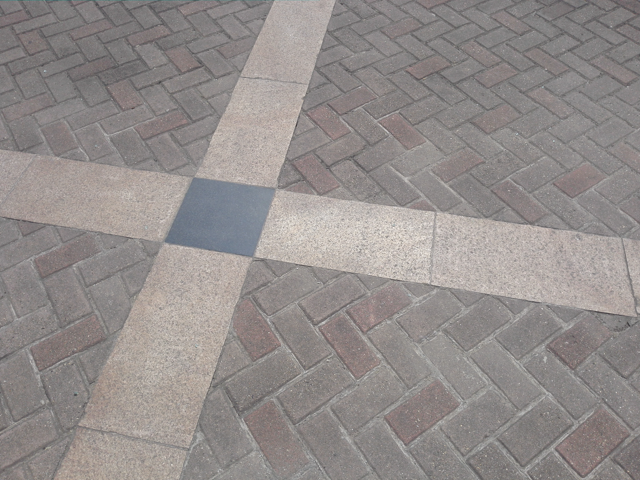} \\
    \hspace{0.3cm}{Input} \hspace{2.5cm}{D-Net} \hspace{2.8cm}{D+I-Net} \hspace{1.7cm}{UnShadowNet}  \hspace{2cm}{GT}
     \vspace{0.25 mm}
    
    \caption{
    \textbf{Qualitative results of progressive addition of various components in UnShadowNet.} DeShadower network (D-Net) alone is capable to remove shadow but it fails to match up the illumination level of the shadow-removed area with the shadow-free region. The DeShadower network accurately handles the illumination level when it is trained contrastively with the Illumination network (D+I Net). There remain some visible artifacts due to improper blending that is taken care of by the Refinement network.
    }
    

    \label{fig:shadow_removal_ablation}
    \vspace{-7mm}
\end{figure*}

\subsection{Implementation Details}
The configuration of the generator is adopted from the DenseUNet architecture \cite{li2018single}. Unlike the conventional UNet architecture \cite{ronneberger2015u}, it uses skip connections to facilitate better information sharing among the symmetric layers. For the discriminator, we employ the architecture of the PatchGAN \cite{isola2017image} discriminator that penalizes generated image structure at the scale of patches instead of at the image level. We develop and train all our models using the PyTorch framework. The proposals are trained using Momentum Optimizer with $1\times10^{-4}$ as the base learning rate for the first $75$ epochs, then we apply linear decay for the rest of the epochs. We train the whole model for a total of $200$ epochs. Momentum was set to $0.9$. All the models were trained on a system comprising one NVIDIA GeForce GTX $2080$Ti GPU and the batch size was set to $1$ for all experiments. In the testing phase, shadow-removed outputs are re-sized to $256\times256$ to compare with the ground truth images, as followed in \cite{le2020shadow, le2021physics}. 
We used the shadow detector by Ding \etal \cite{ding2019argan} to extract the shadow masks during the testing phase.

\subsection{Ablation study}
We considered the adjusted ISTD \cite{wang2018stacked} dataset to perform our ablation studies due to its large volume and common usage in most of the recent shadow removal literature. We design an extensive range of experiments on this dataset in both \textit{weakly-supervised} and \textit{fully-supervised} settings to evaluate the efficacy of the proposed several network components of \textit{UnShadowNet} and find out the best configuration of our model.

\textbf{Network components:}
DeShadower network ($\mathcal{D}$) is the basic unit that acts as the overall shadow remover in the proposal. In the weakly-supervised setup, first, we experiment with only $\mathcal{D}$ for shadow removal (D-Net). We then add the Illumination network ($\mathcal{I}$) to include diverse illumination levels on the non-shadow regions in the image. We couple $\mathcal{I}$ with $\mathcal{D}$ in contrastive learning setup (D+I-Net). After shadow removal, the shadow-free region needs refinement for efficient blending with the non-shadow area. Hence we add a Refinement network ($\mathcal{R}$) with $\mathcal{D}$ where L1 loss guides to preserve the structural details (D+R-Net). Next, we consider illumination-guided contrastive learned refinement (D+R-Net) network where we add $\mathcal{I}$ and that becomes D+I+R-Net. Further improvement is achieved when we add contrastive loss in $\mathcal{R}$ which completes the UnShadowNet framework. In the fully-supervised setup, as described earlier, $\mathcal{I}$ is not used. As a result, we present the study of D-Net, D+R-Net, and UnShadowNet respectively.

Table \ref{tab:shadow_network_ablation} summarizes the ablation study of various proposed network components. Improvement in accuracy is observed due to the addition of $\mathcal{I}$ in contrastive learning setup. $\mathcal{R}$ adds further significant benefit when L1 loss is replaced with contrastive loss. The improvements of the proposed components are consistent in both   self-supervised and fully-supervised learning as reported in the same table. All further experiments are performed based on the configuration marked as UnShadowNet.

\textbf{Curriculum learning:}
Curriculum  Learning \cite{bengio2009curriculum} is a type of learning strategy that allows one to feed easy examples to the neural network first and then gradually increase the complexity of the data. This helps to achieve stable convergence of the global optimum. As per Table \ref{tab:ablation_data_aug}, it is observed that the curriculum learning technique provides considerable improvement when applied along with shadow inpainting and data augmentation.

\begin{figure}
    \centering
    
    \includegraphics[width=0.15\textwidth]{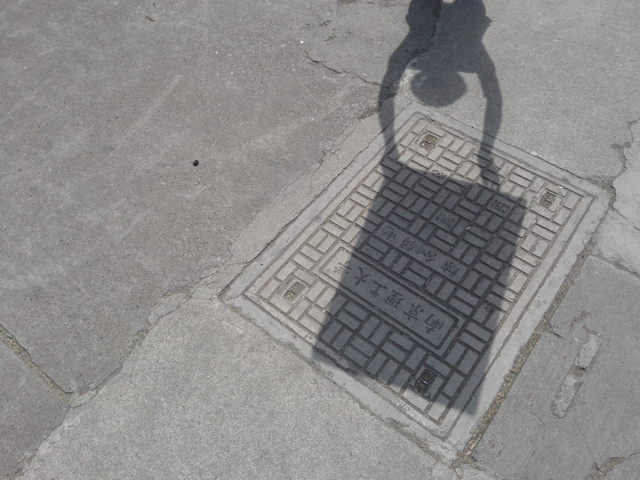}
    \includegraphics[width=0.15\textwidth]{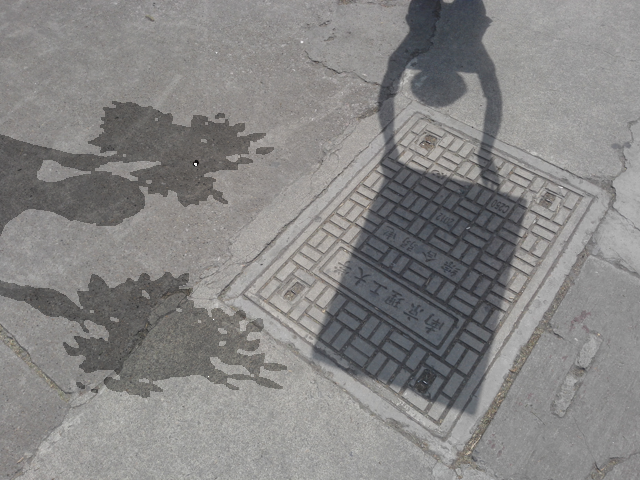}
    \includegraphics[width=0.15\textwidth]{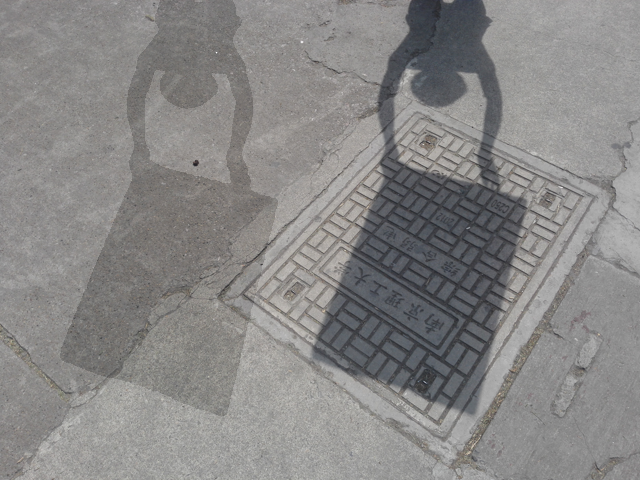}
    
     \vspace{0.25 mm}
    
    \includegraphics[width=0.15\textwidth]{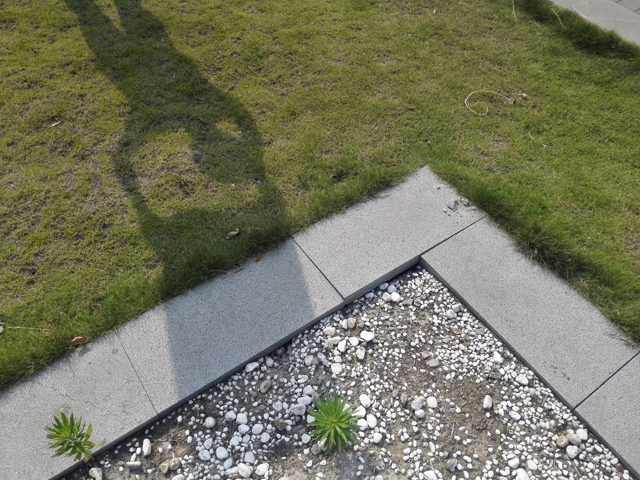}
    \includegraphics[width=0.15\textwidth]{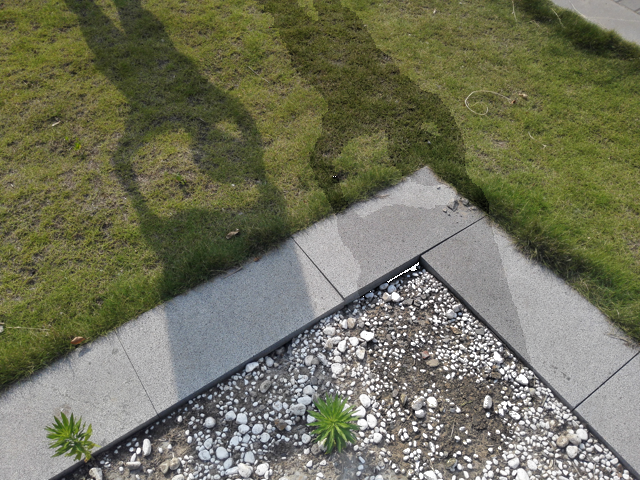}
    \includegraphics[width=0.15\textwidth]{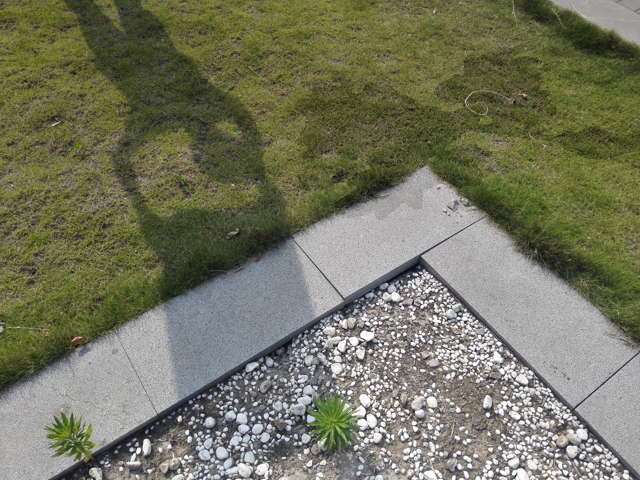}

   \hspace{-1.5cm}{Input} \hspace{1.8cm}{Inpainted Shadow Images} 
    \caption{
    \textbf{Augmented samples with inpainted shadow regions. } 
    }
    \label{fig:shadow_inpainting}
    
    \vspace{-8mm}
    
\end{figure}

\begin{table}[t]
\centering

\scalebox{0.84}{
\begin{tabular}{l|c|c|c|c|c|c} 
\bottomrule

 &  \multicolumn{3}{c|} {\textbf{Shadow Region}} & \multicolumn{3}{c}{\textbf{All}}\\
\cline{1-7}
\cellcolor{blue!25} $\bm{\mu}$  &  \cellcolor{red!25}$\textbf{RMSE}$$\downarrow$  & \cellcolor{red!25}$\textbf{PSNR}$$\uparrow$ & \cellcolor{red!25}$\textbf{SSIM}$$\uparrow$ & \cellcolor{red!25}$\textbf{RMSE}$$\downarrow$  & \cellcolor{red!25}$\textbf{PSNR}$$\uparrow$ & \cellcolor{red!25}$\textbf{SSIM}$$\uparrow$  \\
\cline{1-7}

\midrule
\rowcolor{LightCyan}
\multicolumn{7}{c}{\textbf{Weakly-Supervised Framework}}
\\
\midrule

5  & 24.2 & 27.57 & 0.959 & 19.1 & 24.76 & 0.936\\

25  & 17.1 & 30.13 & 0.974 & 12.4 & 27.30 & 0.942 \\ 

50  & 10.3 & 33.71 & 0.982  & 6.2 & 30.01 & 0.951\\ 

\textbf{75}  & \textbf{8.3} & \textbf{34.47} & \textbf{0.983}  & \textbf{3.8} & \textbf{30.63} & \textbf{0.951} \\ 

100  & 9.8 & 31.56 & 0.980  & 5.6 & 28.52 & 0.946 \\ 

\midrule
\midrule
\rowcolor{LightCyan}
\multicolumn{7}{c}{\textbf{Fully-Supervised Framework}}
\\
\midrule

-15  & 13.2 & 31.57 & 0.979 & 9.7 & 29.26 & 0.949\\

0  & 5.9 & 36.19 & 0.989 & 3.3 & 31.98 & 0.959 \\ 

15  & 6.8 & 35.53 & 0.988 & 4.1 & 31.07 & 0.959 \\ 

30  & 11.6 & 33.87 & 0.983 & 6.3 & 30.56 & 0.952 \\ 

\cline{1-7}

\toprule

\end{tabular}
}

\caption{\textbf{Ablation study on illuminance factor ($\mu$)} of the proposed UnShadowNet in both weakly-supervised and fully-supervised setup on ISTD \cite{wang2018stacked} dataset using RMSE, PSNR, and SSIM metrics.}
\label{tab:mu_table}

\vspace{-7mm}

\end{table}

\begin{table*}[!h]
\centering

\scalebox{0.85}{
\begin{tabular}{l|c|c|c|c|c|c|c|c|c|c} 
\bottomrule

  &  & \multicolumn{3}{c|} {\textbf{Shadow Region}} & \multicolumn{3}{c|}{\textbf{Non-Shadow Region}} & \multicolumn{3}{c}{\textbf{All}}\\
\cline{1-11}
\cellcolor{blue!25} $\textbf{Methods}$  & \cellcolor{blue!25} $\textbf{Training Data}$ & \cellcolor{red!25}$\textbf{RMSE}$$\downarrow$  & \cellcolor{red!25}$\textbf{PSNR}$$\uparrow$ & \cellcolor{red!25}$\textbf{SSIM}$$\uparrow$ & \cellcolor{red!25}$\textbf{RMSE}$$\downarrow$  & \cellcolor{red!25}$\textbf{PSNR}$$\uparrow$ & \cellcolor{red!25}$\textbf{SSIM}$$\uparrow$ & \cellcolor{red!25}$\textbf{RMSE}$$\downarrow$  & \cellcolor{red!25}$\textbf{PSNR}$$\uparrow$ & \cellcolor{red!25}$\textbf{SSIM}$$\uparrow$  \\
\cline{1-11}

Yang \etal \cite{yang2012shadow} & -- & 23.2 & 21.57 & 0.878 & 14.2 & 22.25 & 0.782 & 15.9 & 20.26 & 0.706\\

Gong and Cosker \cite{gong2014interactive}  & -- & 13.0 & 30.53 & 0.972  & 2.6 & 36.63 & 0.982  & 4.3 & 28.96 & 0.943 \\ 
\cline{1-11}
Guo \etal \cite{guo2012paired} & Non.Shd.+Shd (Paired) & 20.1  & 26.89 & 0.960 & 3.1  & 35.48 & 0.975 & 6.1  & 25.51 & 0.924  \\

ST-CGAN \cite{wang2018stacked} & Non.Shd.+Shd (Paired) & 12.0  & 31.70 & 0.979 & 7.9  & 26.39 & 0.956 & 8.6  & 24.75 & 0.927  \\

SP+M-Net* \cite{le2019shadow} & Non.Shd.+Shd (Paired) & 8.1  & 35.08 & 0.984 & 2.8  & 36.38 & 0.979 & 3.6  & 31.89 & 0.953  \\

G2R-ShadowNet \textit{Sup.*} \cite{liu2021shadow}  & Non.Shd.+Shd (Paired) & 7.9  & 36.12 & 0.988 & 2.9  & 35.21 & 0.977 & 3.6  & 31.93 & 0.957  \\
\rowcolor{YellowGreen}
\textbf{UnShadowNet \textit{Sup.*}} & \textbf{Non.Shd.+Shd (Paired)} & \textbf{5.9}  & \textbf{36.19} & \textbf{0.989} & \textbf{2.7}  & \textbf{36.44} & \textbf{0.978} & \textbf{3.3}  & \textbf{31.98} & \textbf{0.959}  \\
\cline{1-11}
Mask-ShadowGAN* \cite{hu2019mask} & Shd.Free(Unpaired) & 10.8  & 32.19 & 0.984 & 3.8  & 33.44 & 0.974 & 4.8  & 28.81 & 0.946  \\
LG-ShadowNet* \cite{liu2021shadow_lg} & Shd.Free(Unpaired) & 9.9  & 32.44 & 0.982 & 3.4  & 33.68 & 0.971 & 4.4  & 29.20 & 0.945  \\
\cline{1-11}
Le \etal* \cite{le2020shadow} & Shd.Mask & 10.4  & 33.09 & 0.983 & 2.9  & 35.26 & 0.977 & 4.0  & 30.12 & 0.950  \\
G2R-ShadowNet* \cite{liu2021shadow} & Shd.Mask & 8.9  & 33.58 & 0.979 & 2.9  & 35.52 & 0.976 & 3.9  & 30.52 & 0.944  \\
\rowcolor{YellowGreen}
\textbf{UnShadowNet}  & \textbf{Shd.Mask} & \textbf{8.3} & \textbf{34.47} & \textbf{0.984} & \textbf{2.9} & \textbf{35.51} & \textbf{0.977} & \textbf{3.8} & \textbf{30.63} & \textbf{0.951}\\
\toprule

\end{tabular}
}

\caption{\textbf{Quantitative comparison of two variants of \textit{UnShadowNet} with other state-of-the-art shadow removal methods} using RMSE, PSNR and SSIM metrics. Methods marked with `*' were evaluated on the adjusted ISTD \cite{wang2018stacked} dataset. Scores of the other methods are computed on the ISTD dataset and obtained from their respective publications.}
\label{tab:shadow_main_table}
\vspace{-3mm}

\end{table*}

\begin{table}[t]
\centering

\scalebox{0.9}{
\begin{tabular}{c|c|c|c}
\bottomrule
\cellcolor{blue!25} $\textbf{Methods}$  & \cellcolor{red!25}$\textbf{Shadow}$$\downarrow$ & \cellcolor{red!25}$\textbf{Non-Shadow}$$\downarrow$ & \cellcolor{red!25}$\textbf{All}$$\downarrow$  \\


\midrule

Guo \etal \cite{guo2011single} & 18.95 & 7.46 & 9.30\\ 

Zhang \etal \cite{zhang2015shadow} & 9.77 & 7.12 & 8.16\\ 

Iizuka \etal \cite{iizuka2017globally} & 13.46 & 7.67 & 8.82 \\ 

Wang \etal \cite{wang2018high} & 10.63 & 6.73 & 7.37 \\

DeshadowNet \cite{qu2017deshadownet} & 12.76 & 7.19 & 7.83\\ 

MaskShadow-GAN* \cite{hu2019mask} & 12.67 & 6.68 & 7.41 \\

ST-CGAN \cite{wang2018stacked} & 10.31 & 6.92 & 7.46\\ 

Cun \etal \cite{cun2020towards} & 11.4 & 7.2 & 7.9 \\ 

AngularGAN \cite{sidorov2019conditional} & 9.78 & 7.67 & 8.16 \\ 

RIS-GAN \cite{zhang2020ris} & 8.99 & 6.33 & 6.95 \\ 

CANet \cite{chen2021canet} & 8.86 & 6.07 & 6.15 \\ 

Hu \etal \cite{hu2019direction} & 7.6 & 3.2 & 3.9 \\ 

Fu \etal \cite{fu2021auto} & 7.77 & 5.56 & 5.92 \\ 

\cline{1-4}
\rowcolor{YellowGreen}
\textbf{UnShadowNet \textit{Sup.}} & \textbf{7.01}  & \textbf{4.58} & \textbf{5.17} \\ 
\rowcolor{YellowGreen}
\textbf{UnShadowNet} & \textbf{9.18}  & \textbf{5.16} & \textbf{6.08} \\


\bottomrule
\end{tabular}
}
\caption{\textbf{Comparative study of fully and weakly-supervised UnShadowNet with other fully supervised state-of-the-art shadow removal methods on ISTD \cite{wang2018stacked} dataset using RMSE metric}. The (*) marked method was trained using unpaired data.
}
\label{tab:istd_full_sup}
\vspace{-2.5mm}

\end{table}

\begin{table}[t]
\centering

\scalebox{0.96}{
\begin{tabular}{c|c|c|c}
\bottomrule
\cellcolor{blue!25} $\textbf{Methods}$  & \cellcolor{red!25}$\textbf{Shadow}$$\downarrow$ & \cellcolor{red!25}$\textbf{Non-Shadow}$$\downarrow$ & \cellcolor{red!25}$\textbf{All}$$\downarrow$  \\


\midrule

Guo \etal \cite{guo2011single} & 22.0 & 3.1 & 6.1\\ 

Gong \etal \cite{gong2016interactive} & 13.3 & -- & --\\ 

ST-CGAN \cite{wang2018stacked} & 13.4 & 7.7 & 8.7\\

DeshadowNet \cite{qu2017deshadownet} & 15.9 & 6.0 & 7.6\\

MaskShadow-GAN* \cite{hu2019mask} & 12.4 & 4.0 & 5.3 \\ 


SP+M-Net \cite{le2021physics} & 7.9 & 3.1 & 3.9 \\ 

Fu \etal \cite{fu2021auto} & 6.5 & 3.8 & 4.2 \\ 

SP+M+I-Net \cite{le2021physics} & 6.0 & 3.1 & 3.6 \\ 

\cline{1-4}
\rowcolor{YellowGreen}
\textbf{UnShadowNet \textit{Sup.}} & \textbf{5.9}  & \textbf{2.7} & \textbf{3.3} \\ 
\rowcolor{YellowGreen}
\textbf{UnShadowNet} & \textbf{8.3}  & \textbf{2.9} & \textbf{3.8} \\ 

\bottomrule
\end{tabular}
}
\caption{\textbf{Comparative study of fully and weakly supervised UnShadowNet with other state-of-the-art shadow removal methods on adjusted ISTD  \cite{le2019shadow} dataset using RMSE metric}. The (*) marked method was trained using unpaired data.
}
\label{tab:istdp_full_sup}
\vspace{-6mm}

\end{table}

\textbf{Shadow inpainting:} Appearance of shadows is a natural phenomenon and yet it is not an easy task to define the strong properties of shadow. This is because it does not have distinguishable shape, size, texture, etc. Hence it becomes important to augment the available shadow samples extensively so that they can be effectively learned by the network.

In this work, we estimate the mean intensity values of the existing shadow region of an image (${I_P}$). Then we randomly select a shadow mask (${S_M}$) from the existing set of shadow samples. The mask (${S_M}$) is inpainted on the shadow-free region of the image (${I_P}$). The pixels that belong to the ${S_M}$ in ${I_P}$ will have brightness adjusted as the earlier computed mean. We do not apply the same mean every time, in order to generate diverse shadow regions, the estimated mean value is adjusted by $\pm 5\%$. The main motivations of this inpainting are two-fold: 1) It is difficult to learn complex shadows when it interacts with diverse light sources and other objects in the scene. The inpainted shadows are standalone and will provide an easier reference sample to another shadow segment in ${I_P}$. 2) It also increases the robustness of the network towards shadow removal by inpainting shadows with  more diverse variations. Figure \ref{fig:shadow_inpainting} shows the proposed shadow inpainting with random shadow masks and different shadow intensities. Table \ref{tab:ablation_data_aug} indicates the significant benefits of inpainting complementing the standard  data augmentation.

\textbf{Data augmentation:}
Data augmentation is an essential constituent to regularize any deep neural network-based model. We make use of some of the standard augmentation techniques such as image flipping with a probability of $0.3$, random scaling of images in the range $0.8$ to $1.2$, adding Gaussian noise, blur effect, and enhancing contrast.

Table \ref{tab:ablation_data_aug} sums up the role of curriculum learning, shadow inpainting, and data augmentation individually and the various combinations. This ablation study is performed on both weakly-supervised and fully-supervised setups indicating that both these training strategies are beneficial to learn shadow removal tasks.

\textbf{Illuminance factor ($\bm{\mu}$):}
The DeShadower Network maximizes the information with ``bright" synthetic augmentations generated by the Illumination Network. The effectiveness of the Illumination Network is verified from the results in Fig. \ref{fig:shadow_removal_ablation}. To train the Illumination Network, we sample shadow regions from the dataset and vary their brightness by $\mu-5, \mu, \mu+5$. The different values experimented for the Illuminance factor ($\mu$) are presented in Table \ref{tab:mu_table}. We find that setting the value of $\mu$ at $50$ gives the most optimal results in shadow removal performance. For the fully-supervised setup, since the ground-truth images are available, the optimal level of brightness is obtained from those samples itself, consequently, $\mu = 0$ gives the best performance.

\begin{table}[t]
\centering

\scalebox{1}{
\begin{tabular}{c|c|c|c}
\bottomrule
\cellcolor{blue!25} $\textbf{Methods}$  &
\cellcolor{red!25}$\textbf{Shadow}$$\downarrow$ & \cellcolor{red!25}$\textbf{Non-Shadow}$$\downarrow$ & \cellcolor{red!25}$\textbf{All}$$\downarrow$  \\


\midrule

Guo \etal \cite{guo2011single} & 31.06 & 6.47 & 12.60\\ 

Zhang \etal \cite{zhang2015shadow} & 9.50 & 6.90 & 7.24\\ 

Iizuka \etal \cite{iizuka2017globally} & 19.56 & 8.17 & 16.33 \\ 

Wang \etal \cite{wang2018high} & 17.33 & 7.79 & 12.58 \\ 

DeshadowNet \cite{qu2017deshadownet} & 17.96 & 6.53 & 8.47\\ 

ST-CGAN \cite{wang2018stacked} & 18.64 & 6.37 & 8.23\\ 

Hu \etal \cite{hu2019direction} & 11.31 & 6.72 & 7.83 \\ 

AngularGAN \cite{sidorov2019conditional} & 17.63 & 7.83 & 15.97 \\ 

Cun \etal \cite{cun2020towards} & 8.94 & 4.80 & 5.67 \\ 

Fu \etal \cite{fu2021auto} & 8.56 & 5.75 & 6.51 \\ 

RIS-GAN \cite{zhang2020ris} & 8.22 & 6.05 & 6.78 \\ 

CANet \cite{chen2021canet} & 7.82 & 5.88 & 5.98 \\ 

\cline{1-4}
\rowcolor{YellowGreen}
\textbf{UnShadowNet \textit{Sup.}} & \textbf{7.78}  & \textbf{5.31} & \textbf{5.74} \\ 
\rowcolor{YellowGreen}
\textbf{UnShadowNet} & \textbf{8.92}  & \textbf{5.96} & \textbf{6.61} \\ 

\bottomrule
\end{tabular}
}
\caption{\textbf{Comparative study of fully and weakly-supervised UnShadowNet with other fully-supervised state-of-the-art shadow removal methods on SRD \cite{qu2017deshadownet} dataset using RMSE metric}. No other prior art was found to remove shadows in a weakly supervised fashion on the same dataset.
}
\label{tab:srtd_full_sup}
\vspace{-4mm}

\end{table}

\subsection{Quantitative study}
We evaluate our proposals and compare quantitatively with the state-of-the-art shadow removal techniques on ISTD \cite{wang2018stacked}, Adjusted ISTD \cite{le2019shadow}, and SRD \cite{qu2017deshadownet} benchmark datasets.

\begin{figure*}[t]
    \centering
    \includegraphics[width=0.161\textwidth]{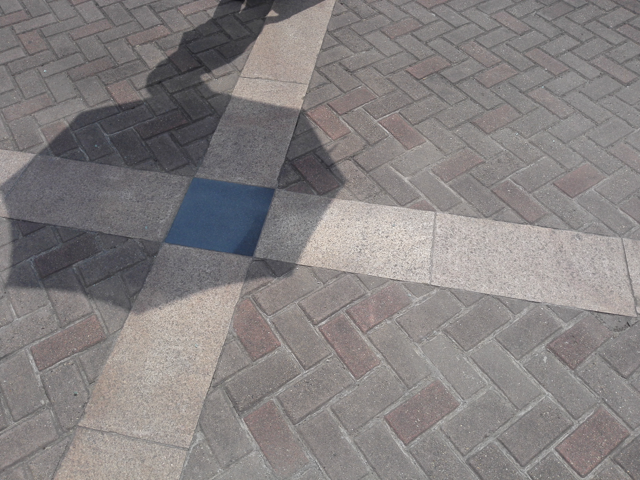}
    \includegraphics[width=0.161\textwidth]{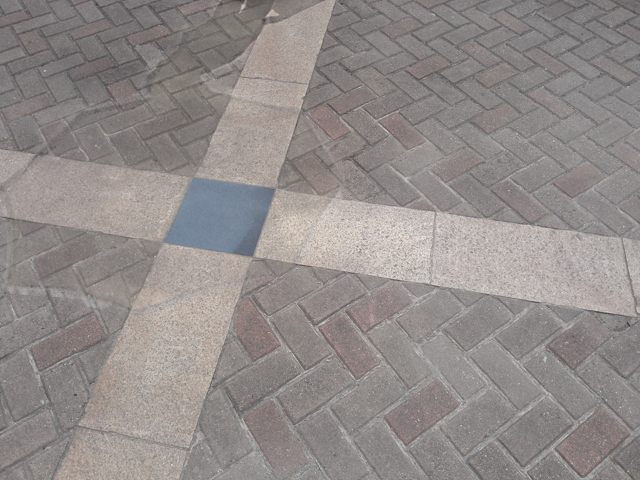}
    \includegraphics[width=0.161\textwidth]{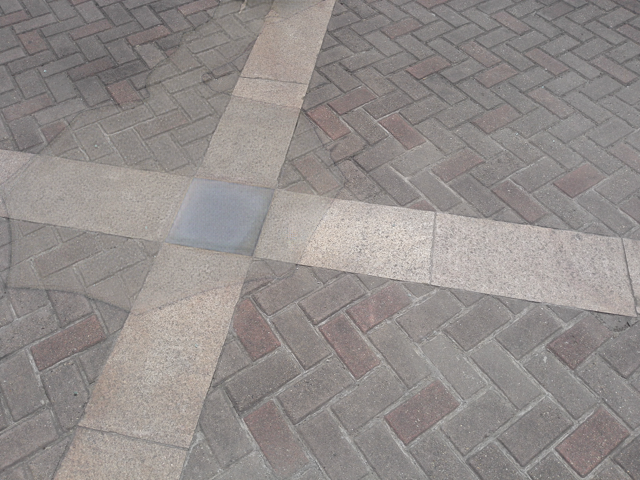}
    \includegraphics[width=0.161\textwidth]{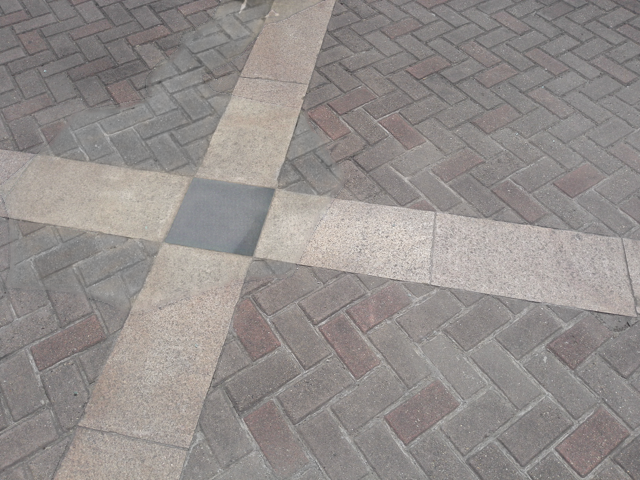}
    \includegraphics[width=0.161\textwidth]{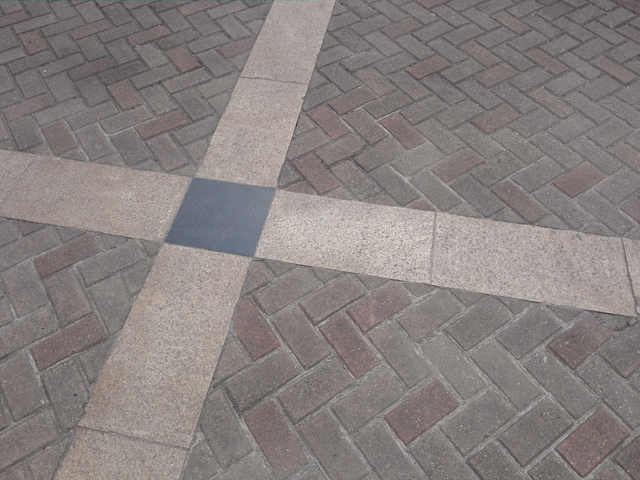}
    \vspace{0.25 mm}\\
    \includegraphics[width=0.161\textwidth]{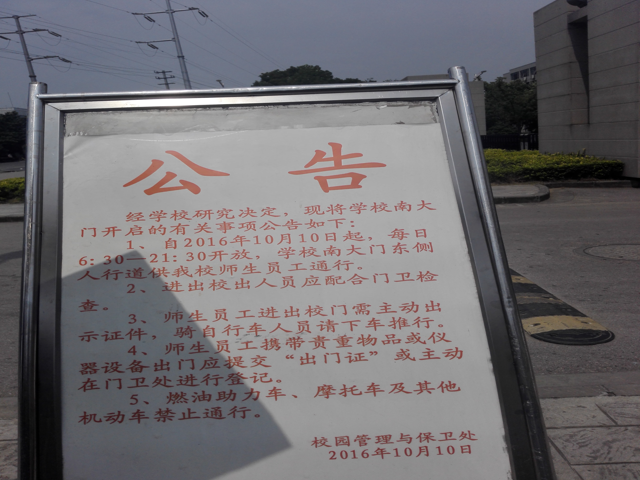}
    \includegraphics[width=0.161\textwidth]{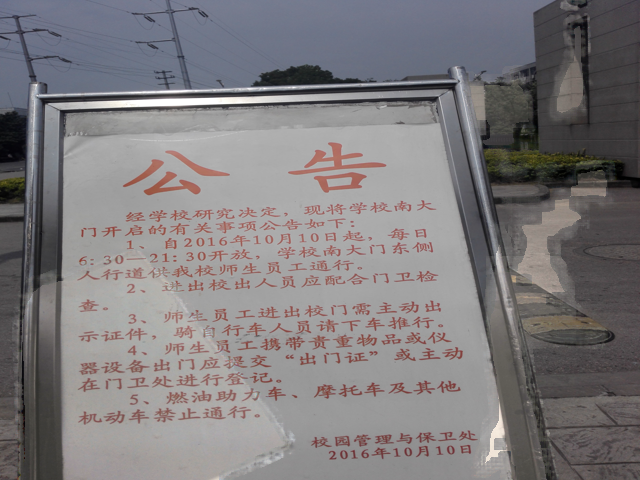}
    \includegraphics[width=0.161\textwidth]{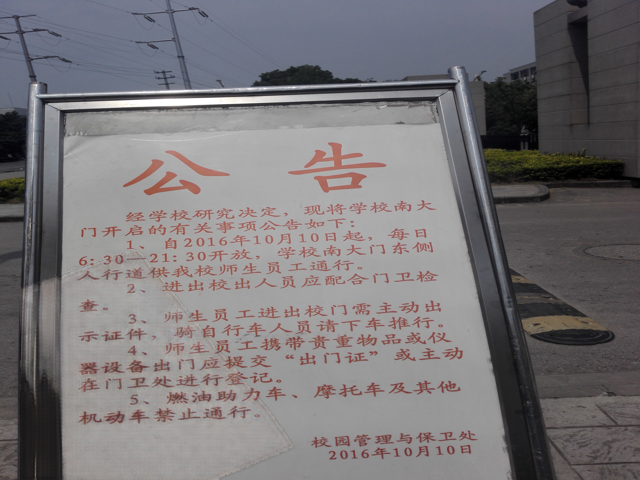}
    \includegraphics[width=0.161\textwidth]{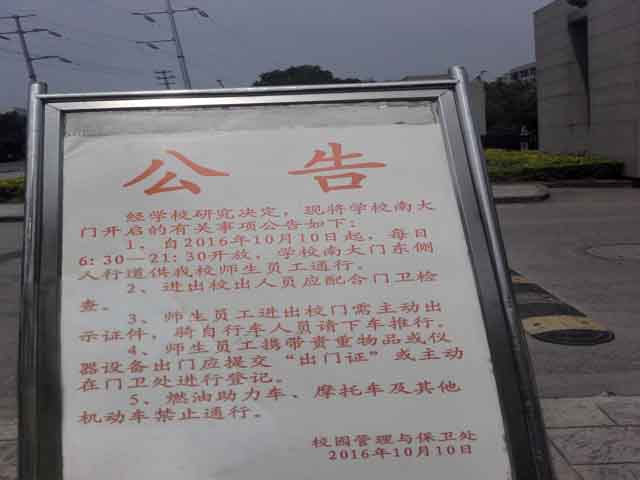}
    \includegraphics[width=0.161\textwidth]{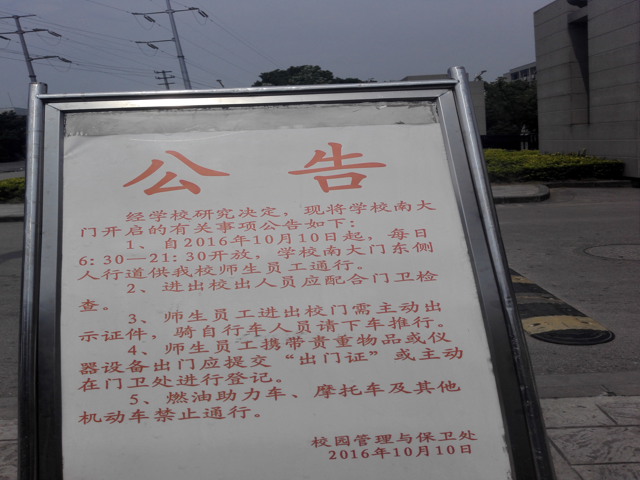}    
    \vspace{0.25 mm}\\
    \includegraphics[width=0.161\textwidth]{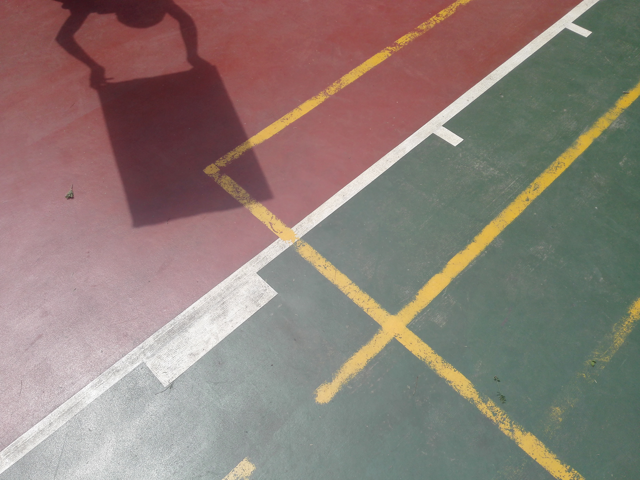}
    \includegraphics[width=0.161\textwidth]{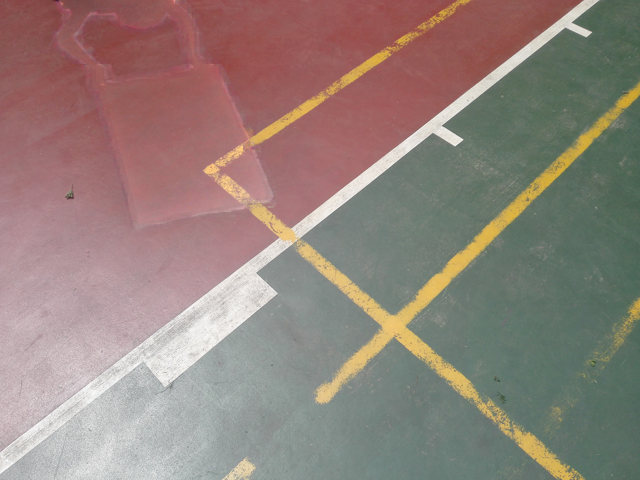}
    \includegraphics[width=0.161\textwidth]{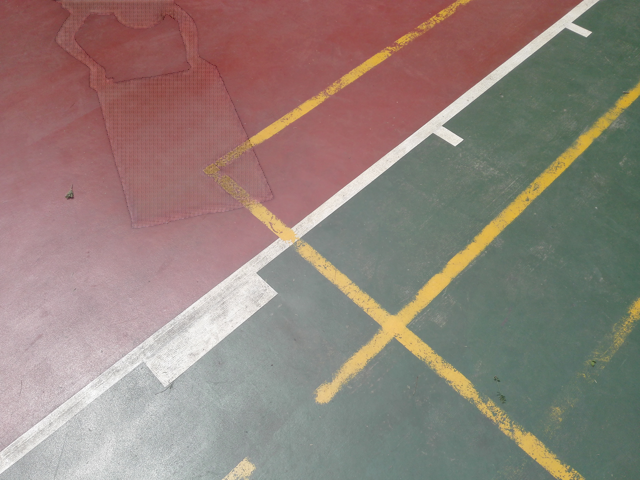}
    \includegraphics[width=0.161\textwidth]{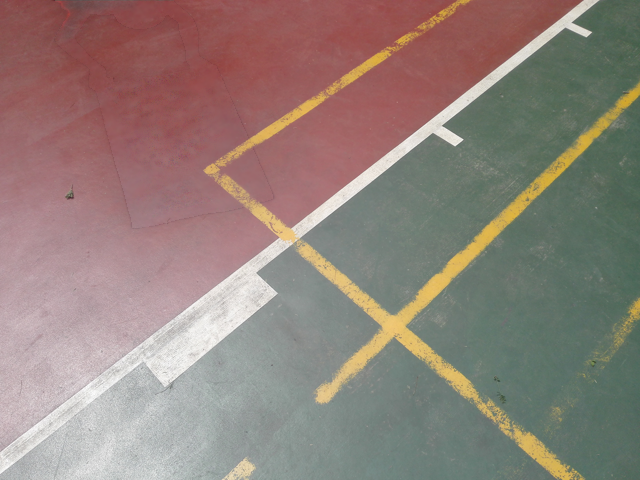}
    \includegraphics[width=0.161\textwidth]{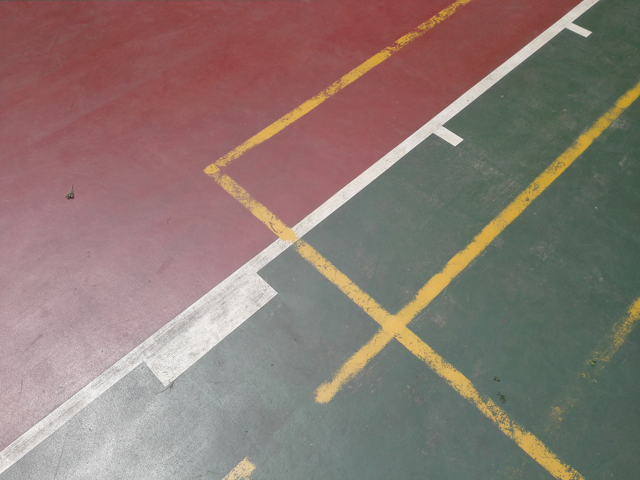}
    \vspace{0.25 mm}\\
    \hspace{-0.25cm}
    \includegraphics[width=0.161\textwidth]{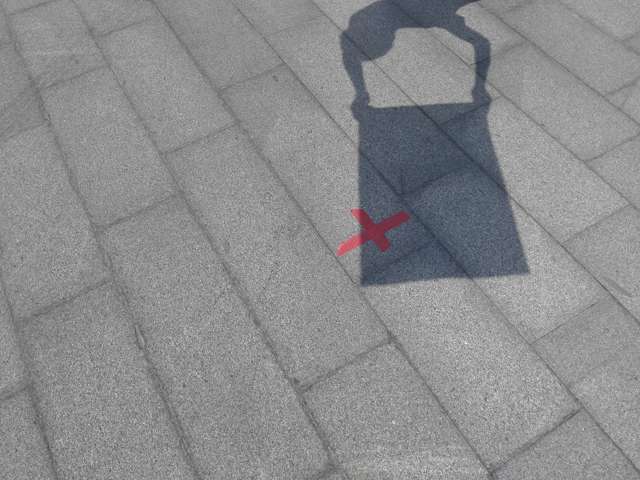}
    \includegraphics[width=0.161\textwidth]{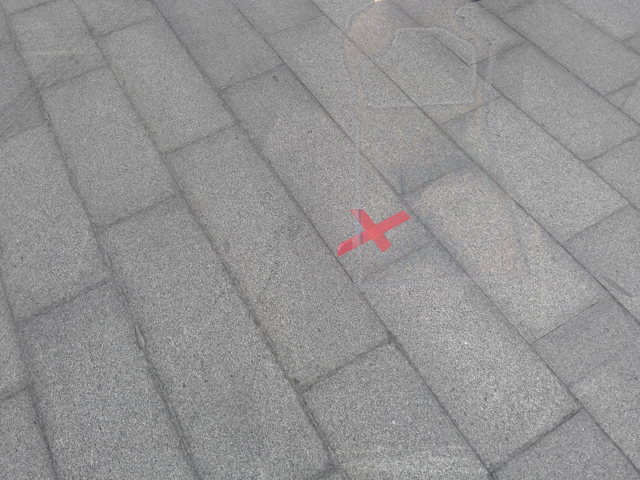}
    \includegraphics[width=0.161\textwidth]{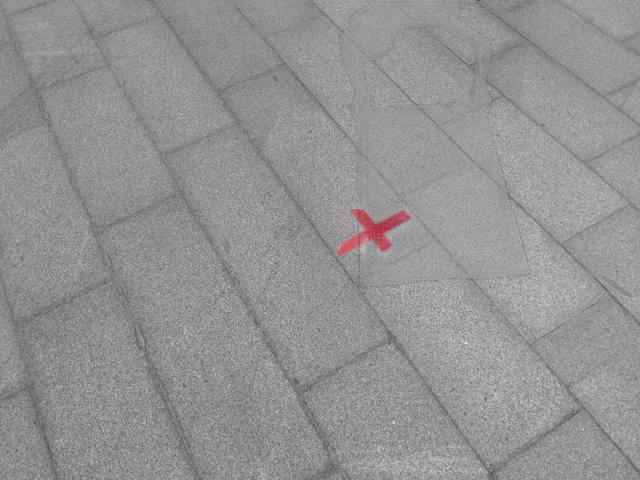}
    \includegraphics[width=0.161\textwidth]{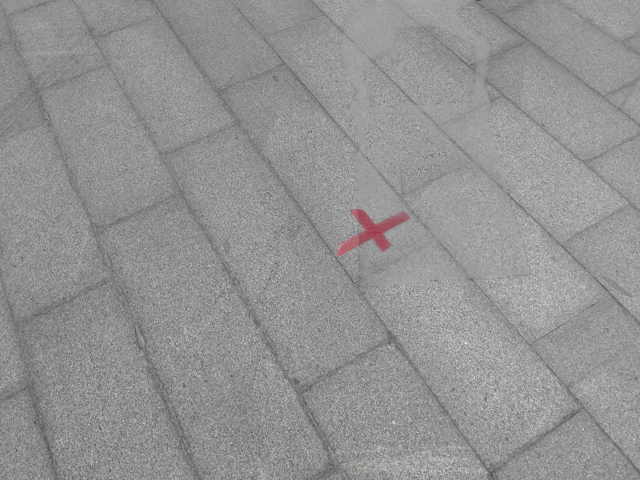}
    \includegraphics[width=0.161\textwidth]{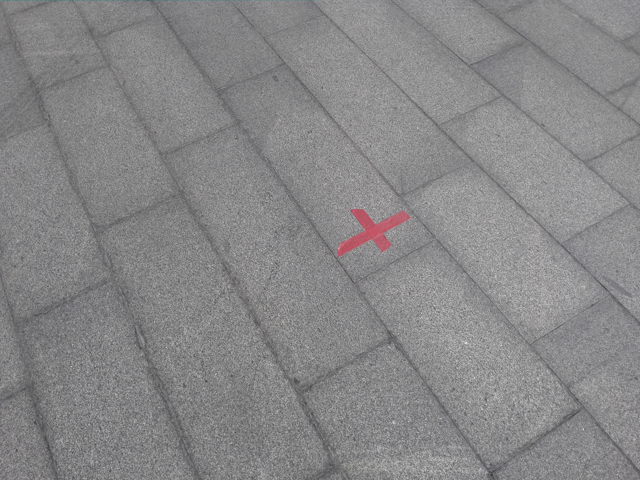} \\
    \hspace{-0.8cm}{Input} \hspace{1.4cm}{Le \etal \cite{le2020shadow}} \hspace{0.6cm}{G2R-ShadowNet \cite{liu2021shadow}} \hspace{0.8cm}{Ours}  \hspace{2cm}{GT}
    \caption{
    \textbf{Qualitative comparison of our proposed method} with other state-of-the-art shadow removal methods that use shadow mask and shadow image as input, on four challenging samples from ISTD \cite{wang2018stacked} dataset.
    }
    

    \label{fig:globfig}
\end{figure*}

\textbf{ISTD:} Table \ref{tab:shadow_main_table} compares the proposed method with the state-of-the-art shadow removal approaches using RMSE, PSNR, and SSIM metrics for shadow, shadow-free, and all regions. We achieve state-of-the-art results and the improvement with respect to all metrics for shadow area in both training setups, namely weakly-supervised (UnshadowNet) and fully-supervised (UnshadowNet \textit{Sup.}), are quite significant. There are a few other fully-supervised shadow removal methods evaluated on ISTD \cite{wang2018stacked} dataset, which we compared with our proposed fully-supervised setup. In this setup as well, as per Table \ref{tab:istd_full_sup}, our proposed method outperforms other state-of-the-art approaches.

\textbf{ISTD+:} Table \ref{tab:istdp_full_sup} shows the performance of our proposed shadow remover on the adjusted ISTD  \cite{le2019shadow} dataset using RMSE metric. The comparison of our method in a fully-supervised setup with other techniques trained in the same fashion demonstrates the robustness of our framework as it shows incremental improvement over the most recent state-of-the-art methods. In addition, we have performed experiments using a weakly-supervised setup where the metrics are  comparable and only slightly behind the fully-supervised model.

\textbf{SRD:} We report and compare our shadow removal results in both the constrained and unconstrained setups with existing fully-supervised methods on SRD \cite{qu2017deshadownet} using RMSE metric. Table \ref{tab:srtd_full_sup} indicates that our proposal trained in a fully-supervised fashion obtains the lowest RMSE in all regions and outperforms the most recent state-of-the-art methods \cite{fu2021auto, zhang2020ris}.

\subsection{Qualitative study}

\begin{figure}
    \centering
    
    \includegraphics[width=0.15\textwidth]{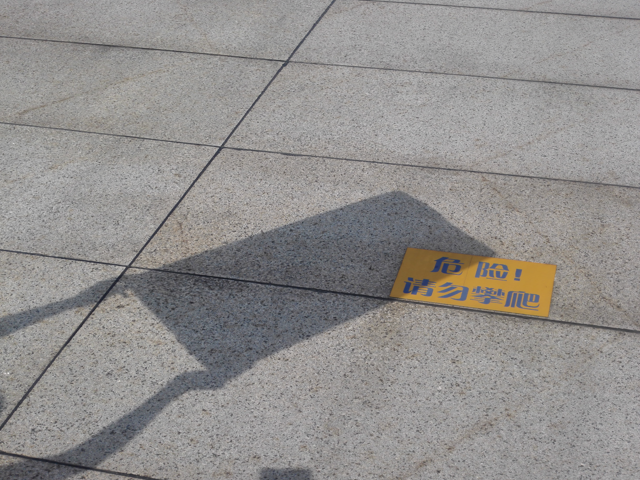}
    \includegraphics[width=0.15\textwidth]{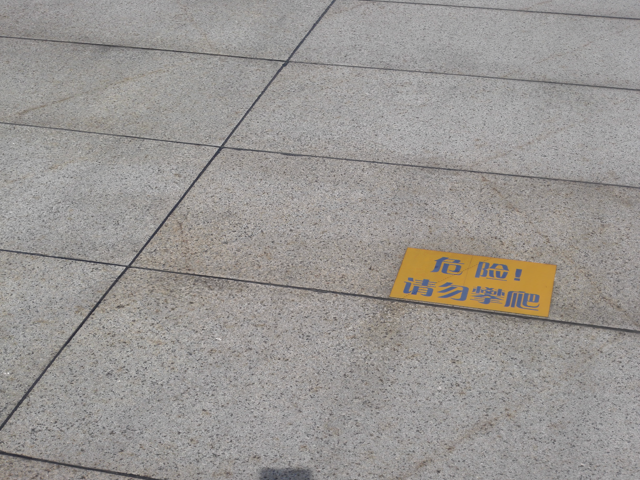}
    \includegraphics[width=0.15\textwidth]{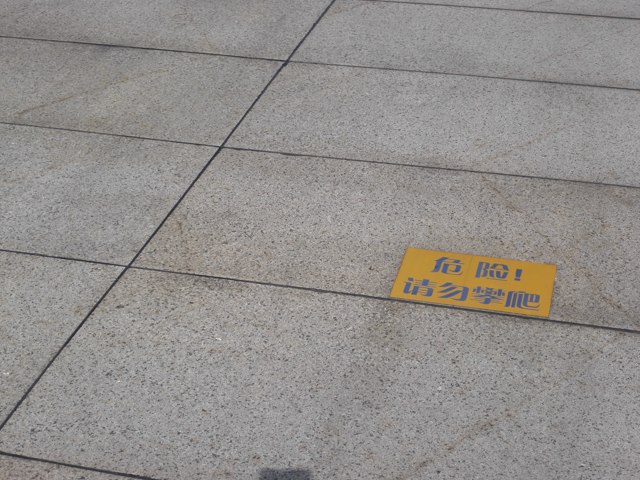}
    
     \vspace{0.4 mm}
    
    \includegraphics[width=0.15\textwidth]{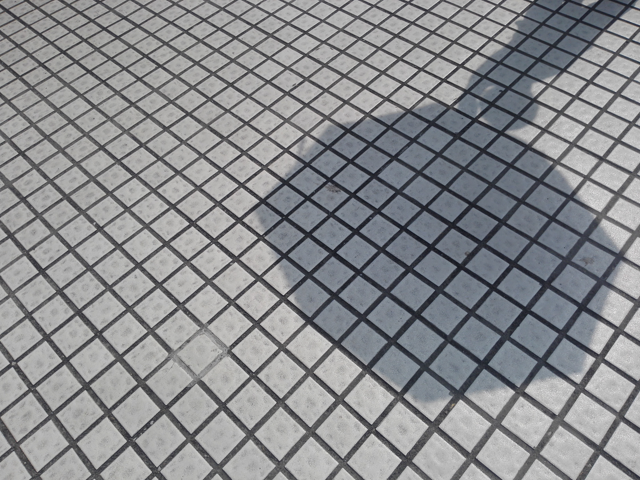}
    \includegraphics[width=0.15\textwidth]{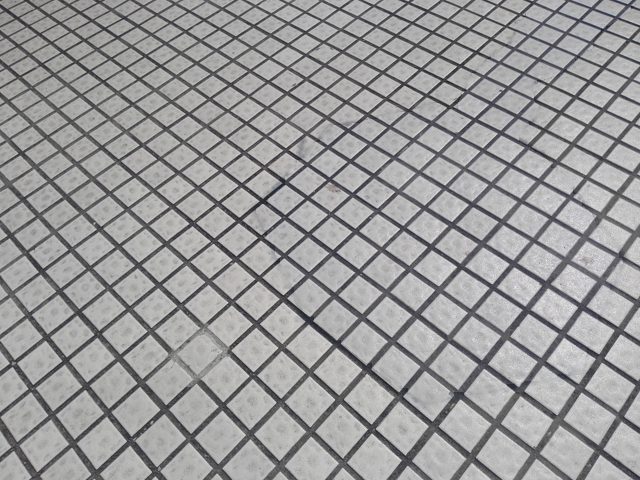}
    \includegraphics[width=0.15\textwidth]{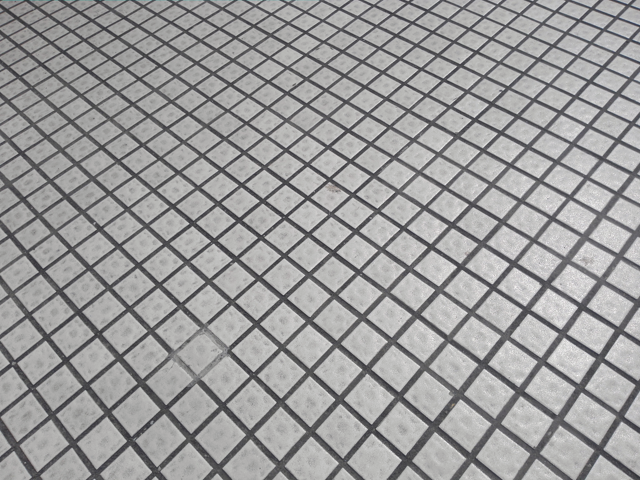}
   
   \hspace{0.6cm}{Input} \hspace{1.4cm}{UnShadowNet} \hspace{0.8cm}{Ground Truth}
   
    \caption{
    \textbf{Qualitative results on the ISTD \cite{wang2018stacked} dataset using \textit{fully-supervised} UnShadowNet setup.}
    }
 \vspace{-6 mm}
 
    \label{fig:results_fully_supervised}
\end{figure}

\begin{figure}[!t]
    \centering
    
    \includegraphics[width=0.15\textwidth]{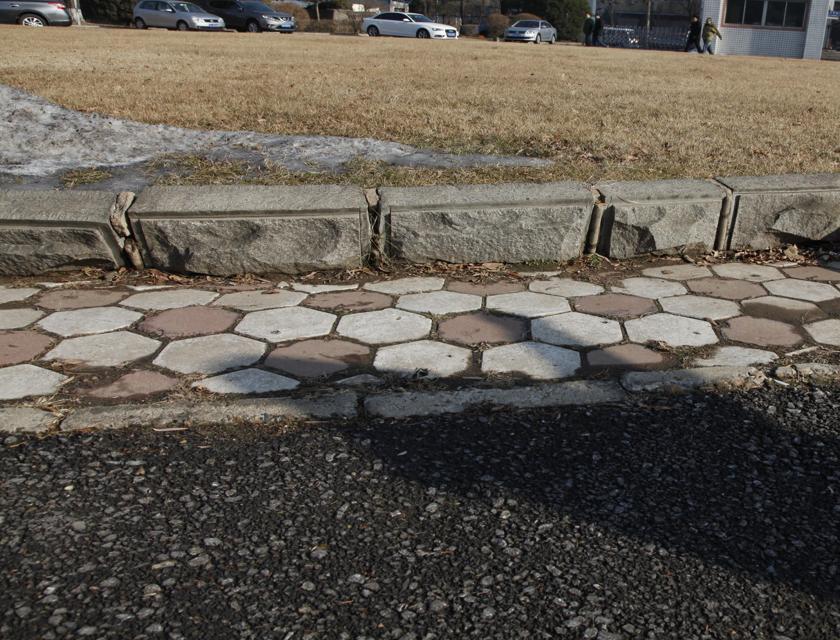}
    \includegraphics[width=0.15\textwidth]{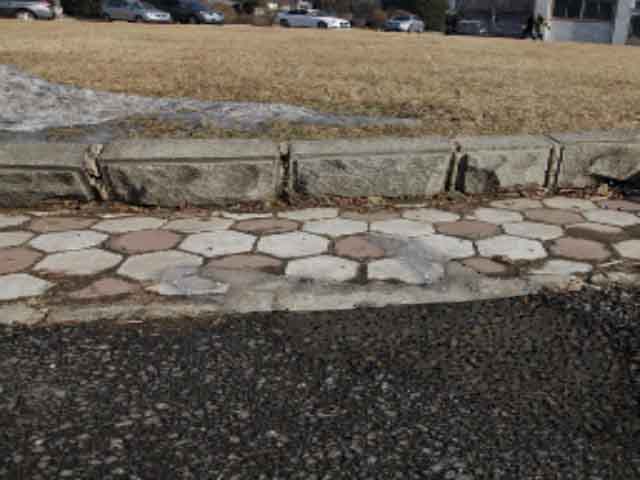}
    \includegraphics[width=0.15\textwidth]{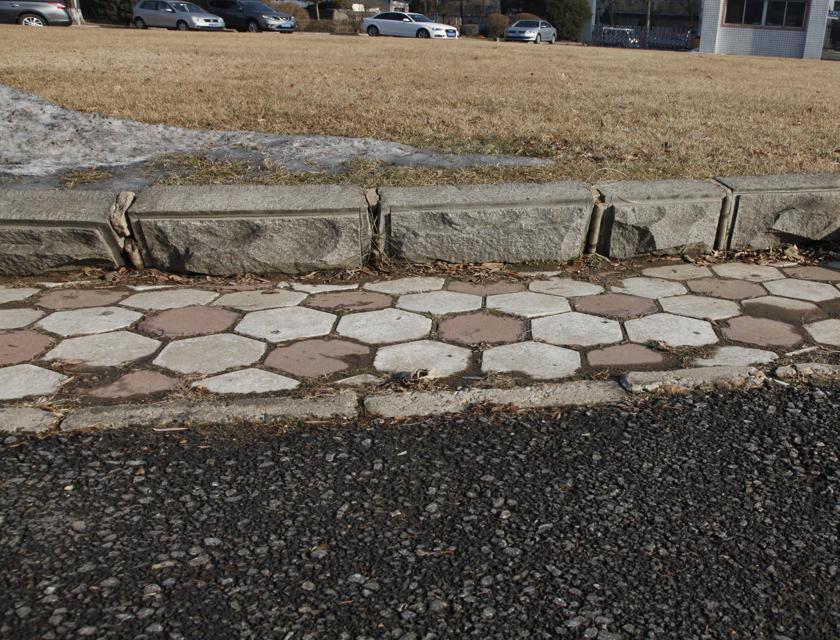}
    
     \vspace{0.25 mm}
    
    \includegraphics[width=0.15\textwidth]{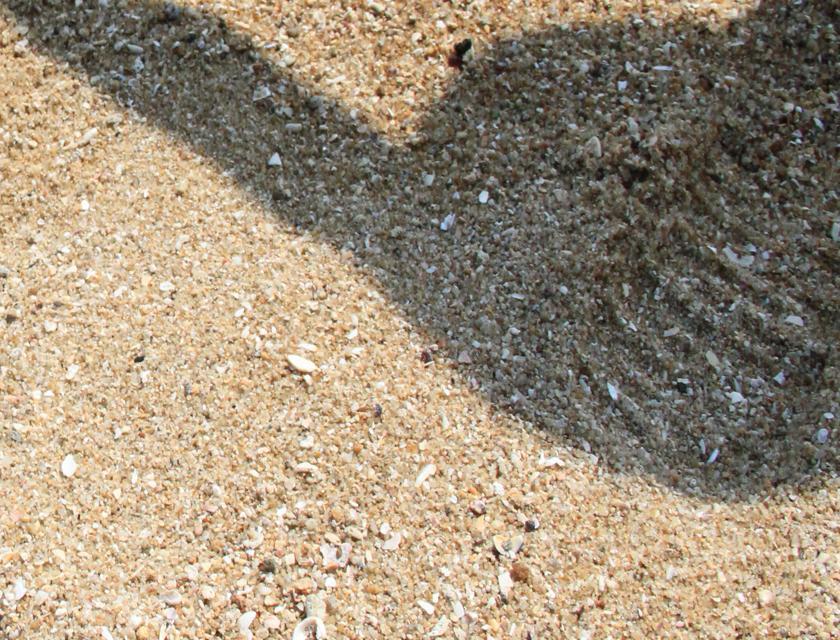}
    \includegraphics[width=0.15\textwidth]{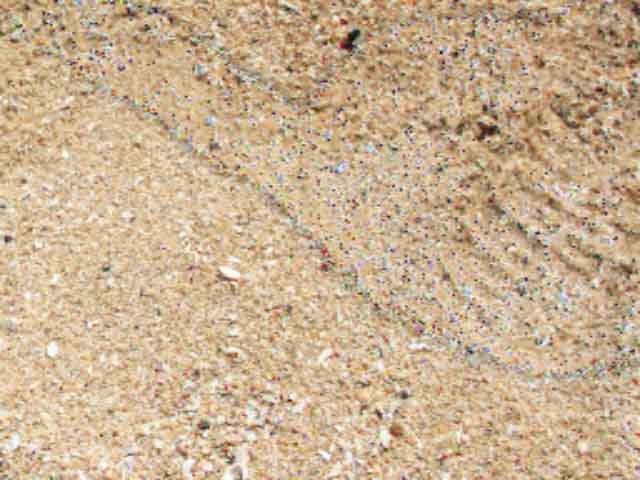}
    \includegraphics[width=0.15\textwidth]{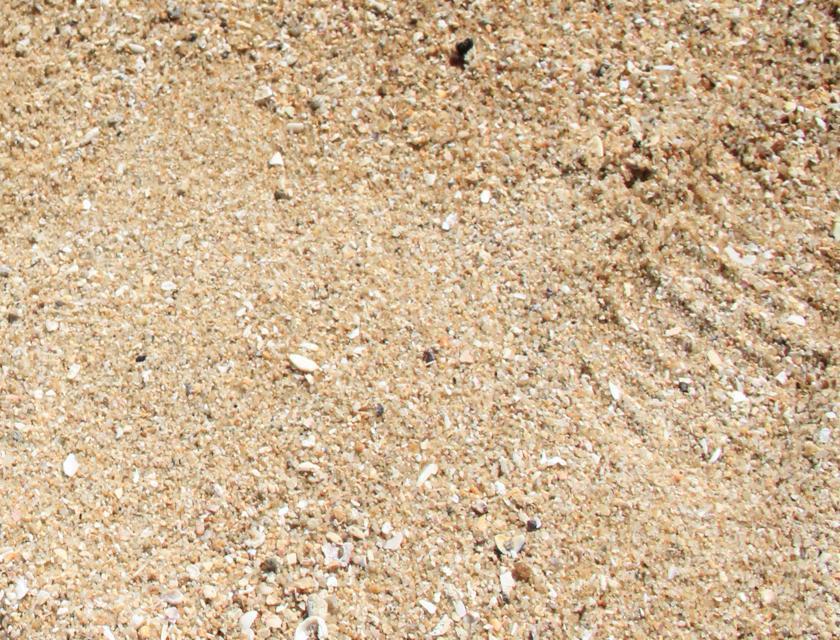}
   \hspace{1.6cm}{Input} \hspace{1.5cm}{UnShadowNet} 
   \hspace{0.4cm} {Ground Truth}

    \caption{
    \textbf{Qualitative results on the SRD \cite{qu2017deshadownet} dataset using \textit{weakly-supervised} UnShadowNet setup.}
    }
 \vspace{-4 mm}
 
    \label{fig:results_srd_self_supervised}
\end{figure}

\begin{figure}
    \centering
    
    \includegraphics[width=0.15\textwidth]{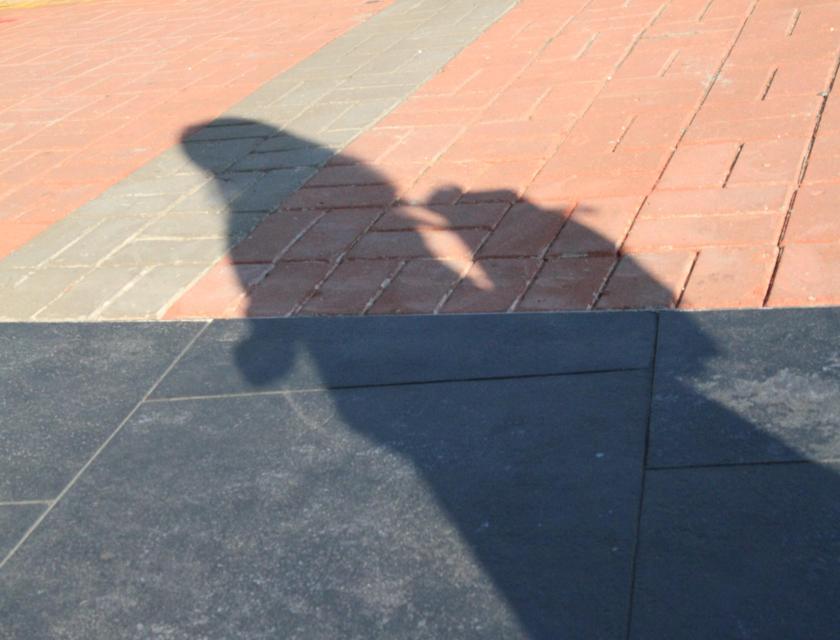}
    \includegraphics[width=0.15\textwidth]{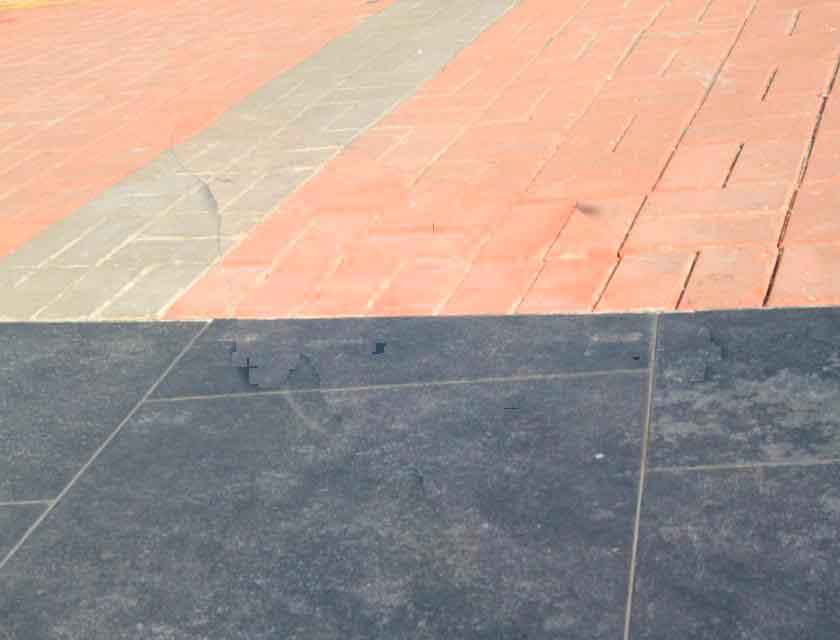}
    \includegraphics[width=0.15\textwidth]{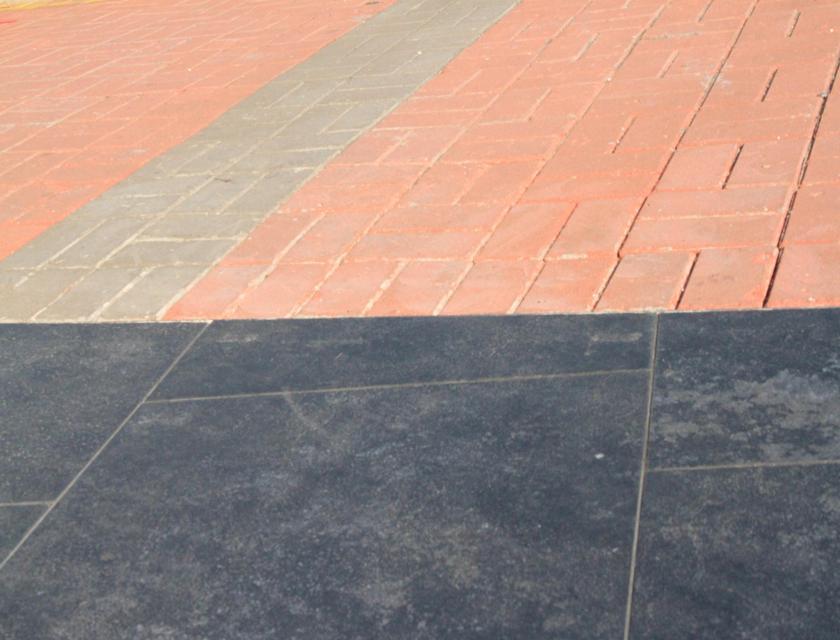}
    
     \vspace{0.25 mm}
    
    \includegraphics[width=0.15\textwidth]{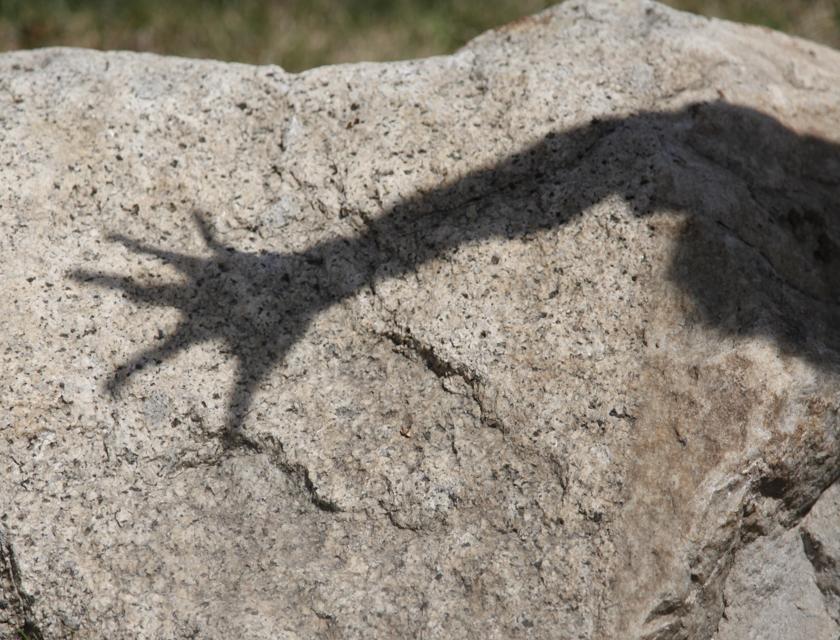}
    \includegraphics[width=0.15\textwidth]{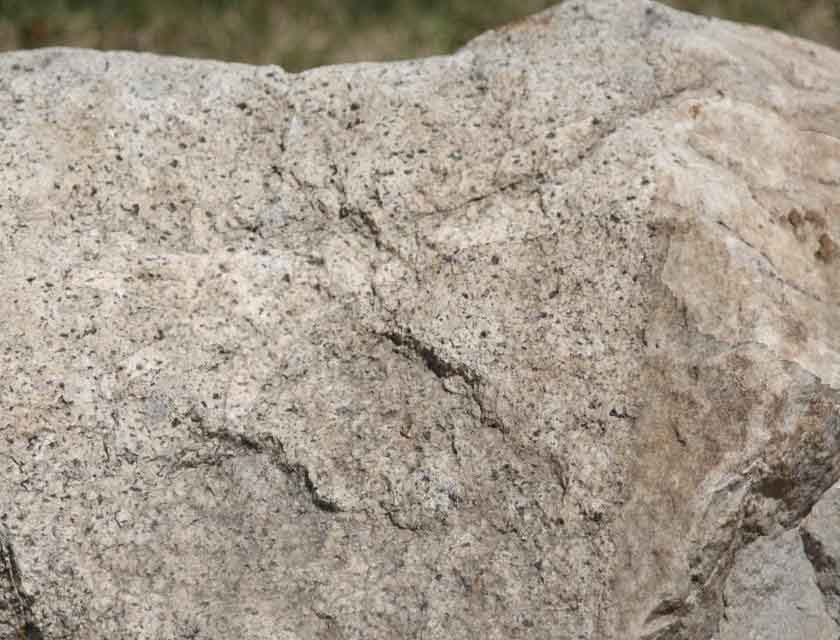}
    \includegraphics[width=0.15\textwidth]{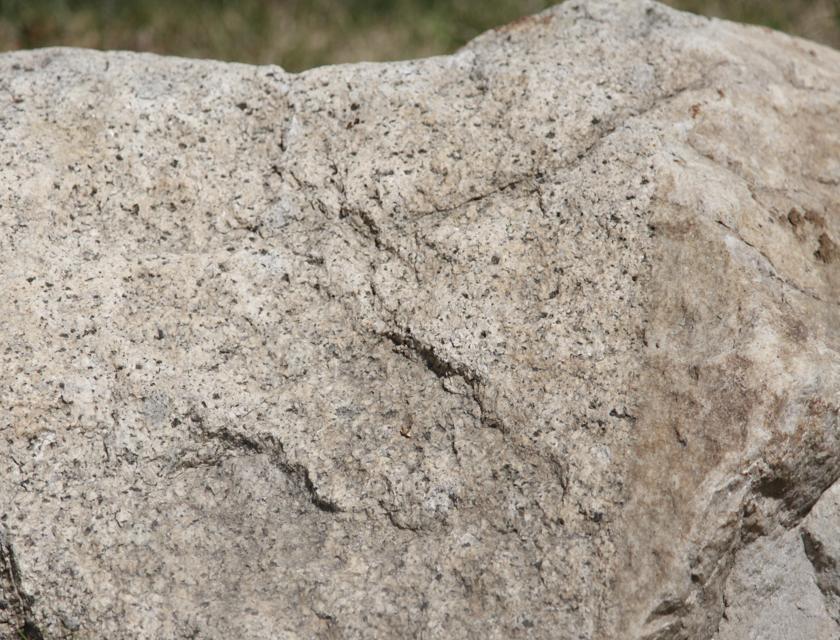}
   \hspace{1.6cm}{Input} \hspace{1.5cm}{UnShadowNet} 
   \hspace{0.4cm} {Ground Truth}

    \caption{
    \textbf{Qualitative results on the SRD \cite{qu2017deshadownet} dataset using \textit{fully-supervised} UnShadowNet setup.}    }
 \vspace{-5 mm}
 
    \label{fig:results_srd_fully_supervised}
\end{figure}

Figure \ref{fig:globfig} shows qualitative results of the proposed model trained in weakly-supervised format on a total of three challenging samples from the ISTD \cite{wang2018stacked} dataset. We also visually compare with two existing and most recently published weakly-supervised shadow removal methods by Le \etal  \cite{le2020shadow} and G2R-ShadowNet \cite{liu2021shadow} respectively. It is clearly observed that UnShadowNet is accurate while removing shadows in complex backgrounds. In addition to the unconstrained setup, Figure \ref{fig:results_fully_supervised} shows the results of our UnshadowNet \textit{Sup.} model on ISTD \cite{wang2018stacked} dataset. It is to be noted that the visual results are not shown on the adjusted ISTD \cite{le2019shadow} dataset because the test samples are the same as in the ISTD dataset, the only difference is in the color of the ground truth. In addition, we consider SRD \cite{qu2017deshadownet} dataset and this is the first work where visual results are presented on the samples from the same dataset. Figure \ref{fig:results_srd_self_supervised} and \ref{fig:results_srd_fully_supervised} demonstrate the results of UnShadowNet in weakly-supervised and fully-supervised setup.

\begin{figure}[!t]
    \centering
    
    \includegraphics[width=0.15\textwidth]{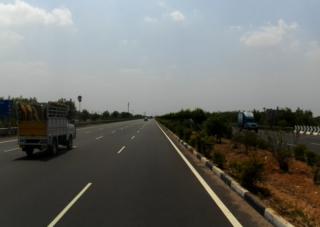}
    \includegraphics[width=0.15\textwidth]{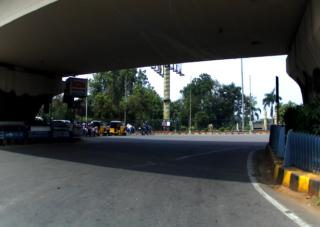}
    \includegraphics[width=0.15\textwidth]{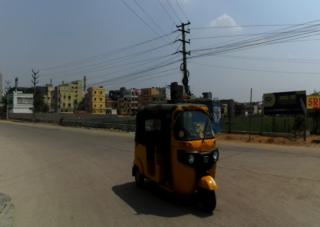}
    
     \vspace{0.25 mm}
    
    \includegraphics[width=0.15\textwidth]{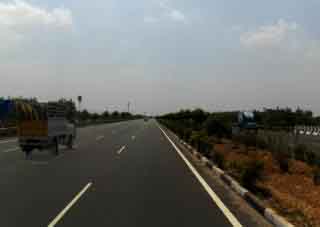}
    \includegraphics[width=0.15\textwidth]{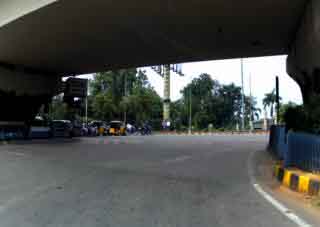}
    \includegraphics[width=0.15\textwidth]{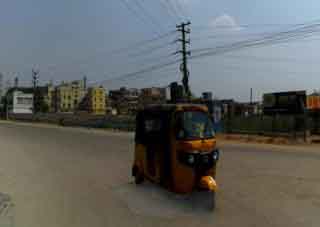}
    \caption{
    \textbf{Qualitative results (\textit{bottom}) on a few input samples (\textit{top}) from IDD dataset \cite{varma2019idd}.} \textit{UnShadowNet} trained on ISTD \cite{wang2018stacked} dataset enables to remove shadow reasonably in automotive scenes. }
 \vspace{-5 mm}
 
    \label{fig:results_automotive}
\end{figure}

\subsection{Runtime Analysis} 
We compare the runtime performance of our model with recent other contemporary architectures. For this purpose, the available code bases were used to estimate the run-time. During inference, LG-ShadowNet \cite{liu2021shadow_lg} takes 0.874 seconds, G2R-ShadowNet \cite{liu2021shadow} takes 0.805 seconds and UnShadowNet takes 0.822 seconds.

\subsection{Evaluation of generalization in an unconstrained automotive dataset}
Automotive object detection and segmentation datasets do not provide shadow labels; thus, it is impossible to quantitatively evaluate these datasets extensively. We sampled a few shadow scenes from the challenging IDD dataset \cite{varma2019idd} which contains varied lighting condition scenes on Indian roads. It was impossible to train our model as shadow masks were unavailable. Thus we used this dataset to evaluate the robustness and generalization of our pre-trained model on novel scenes. The qualitative results are illustrated in Figure \ref{fig:results_automotive}. 
Although the performance of the proposed shadow removal framework is either comparable to the state-of-the-art or superior, it is still not robust to be used in real-world autonomous driving systems. We feel that more extensive datasets have to be built for shadows to perform more detailed studies and we hope this work encourages the creation of these datasets or annotations of shadows in existing datasets.

\section{Limitations and Future Directions}  \label{sec:limit}
As presented in our experiments, UnShadowNet outperforms the existing state-of-the-art in several standard shadow removal datasets. However, there are certain areas that can be improved. Our model relies upon an external shadow detector \cite{ding2019argan} which may not always accurately predict the shadow regions. This may cause resultant areas where the shadow is not removed. In future work, we intend to build a single-stage architecture to incorporate both shadow detection and removal. Since shadows are physical phenomena, another interesting direction would be to exploit the inherent physical properties of illumination that result in shadows.

Moreover, in our research, we observed that the focus is mainly on datasets that have images of a narrow field-of-view and lacks complex situations that may arise in real-life automotive scenes. For future works, we think it will be important to develop a suitable dataset that comprises such challenging scenarios as in real-world automotive settings.

The proposed method is not optimized for run-time and we still obtained a reasonable inference time of 0.822 seconds. With optimization techniques like pruning and multi-task learning, real-time performance can potentially be achieved.

\section{Conclusion}  \label{sec:conc}

In this work, we have developed a novel end-to-end framework consisting of a deep learning architecture for image shadow removal in unconstrained settings. The proposed model can be trained with full or weak supervision. We achieve state-of-the-art results in all the major shadow removal datasets. Although weak supervision has slightly lesser performance, it eliminates the need for shadowless ground truth which is difficult to obtain. To enable the weakly supervised training, we have introduced a novel illumination network which is composed of a generative model used to brighten the shadow region and a discriminator trained using shadow-free patches of the image. It acts as a guide (called illumination critic) for producing illuminated samples by the generator. DeShadower, another component of the proposed framework is trained in a contrastive way with the help of illuminated samples which are generated by the preceding part of the network. Finally, we propose a refinement network that is trained in a contrastive way and is used for fine-tuning the shadow-removed image obtained as an output of the DeShadower. We perform ablation studies to show that the three components of our proposed framework, namely the illuminator, Deshadower, and refinement network work effectively together. To evaluate the generalization capacity of the proposed approach, we tested a few novel samples of shadow-affected images from a generic automotive dataset and obtained promising results of shadow removal. Shadow removal continues to be a challenging problem in dynamic automotive scenes and we hope this work encourages further dataset creation and research in this area.


\section{Acknowledgement}

We would like to thank Valeo for encouraging advanced research. Many thanks to Tuan-Hung Vu (valeo.ai, France), Saikat Roy (DKFZ, Germany) and Aniruddha Saha (University of Maryland, Baltimore County) for providing a detailed review prior to submission. 


\bibliographystyle{IEEEtran}
\bibliography{IEEEfull}


\begin{IEEEbiography}
[{\includegraphics[width=1in,height=1.5in,clip,keepaspectratio]{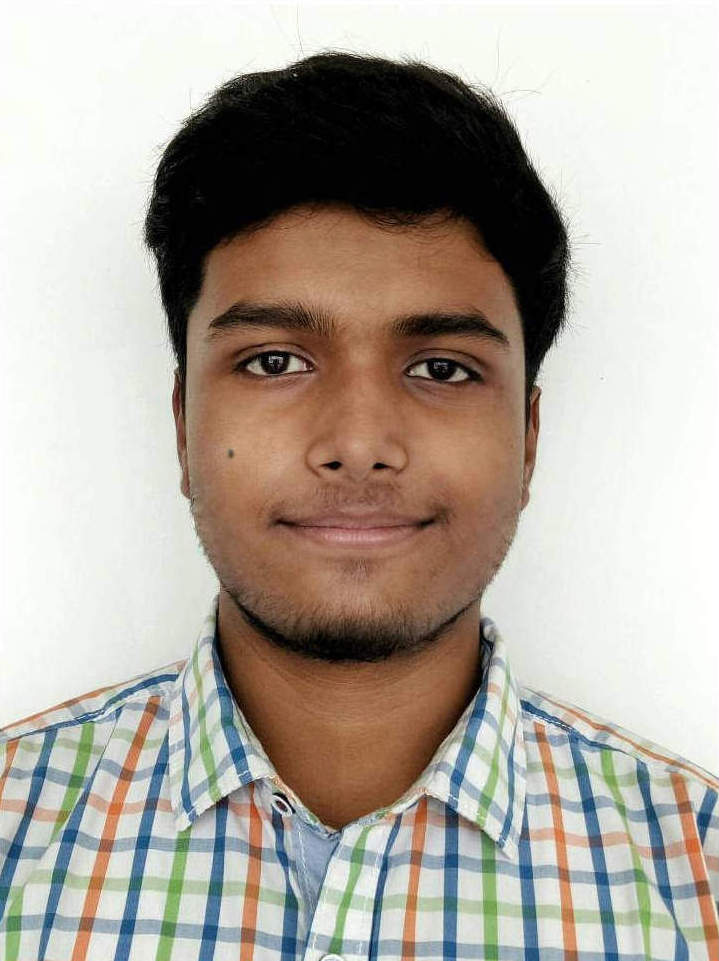}}]{Subhrajyoti Dasgupta} is a Master's student at Mila and Université de Montréal. He completed his Bachelor’s Thesis under Prof. Ujjwal Bhattacharya at Computer Vision and Pattern Recognition Unit at Indian Statistical Institute, Kolkata, and later continued there as a researcher. He has also worked as a Deep Learning Project Trainee at Bhabha Atomic Research Center, Mumbai. His interests mainly comprise areas in Computer Vision like Multi-modal Learning, Scene Understanding, Generative models, and also its applications in inter-disciplinary domains like autonomous driving and climate research.
\end{IEEEbiography}

\begin{IEEEbiography}
[{\includegraphics[width=1in,height=1.5in,clip,keepaspectratio]{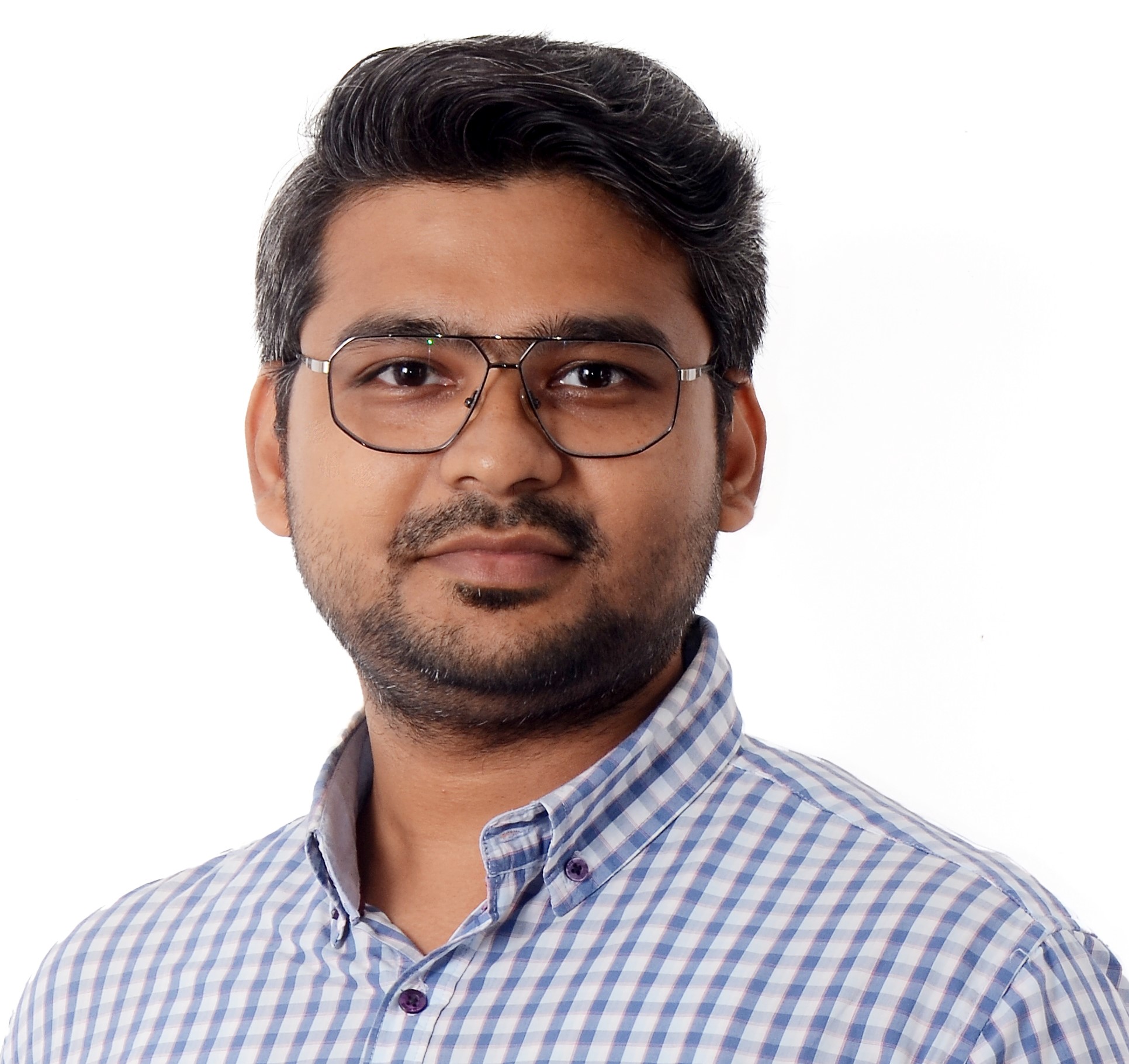}}]{Arindam Das} is an AI Software Architect in the department of Driving Software and Systems (DSW) at Valeo India where he also holds the title of Expert in AI. He is responsible to design AI algorithms to support various features for autonomous driving systems. He is currently pursuing a Ph.D. degree in the Department of Electronic \& Computer Engineering at the University of Limerick, Ireland. He has almost $10$ years of industry experience in computer vision, deep learning \& document analysis. He has authored $17$ peer-reviewed publications \& $48$ patents. His current area of research interest includes weakly-supervised learning, domain adaptation, image restoration, and multimodal learning.
\end{IEEEbiography}

\begin{IEEEbiography}[{\includegraphics[width=1in,height=2.5in,clip,keepaspectratio]{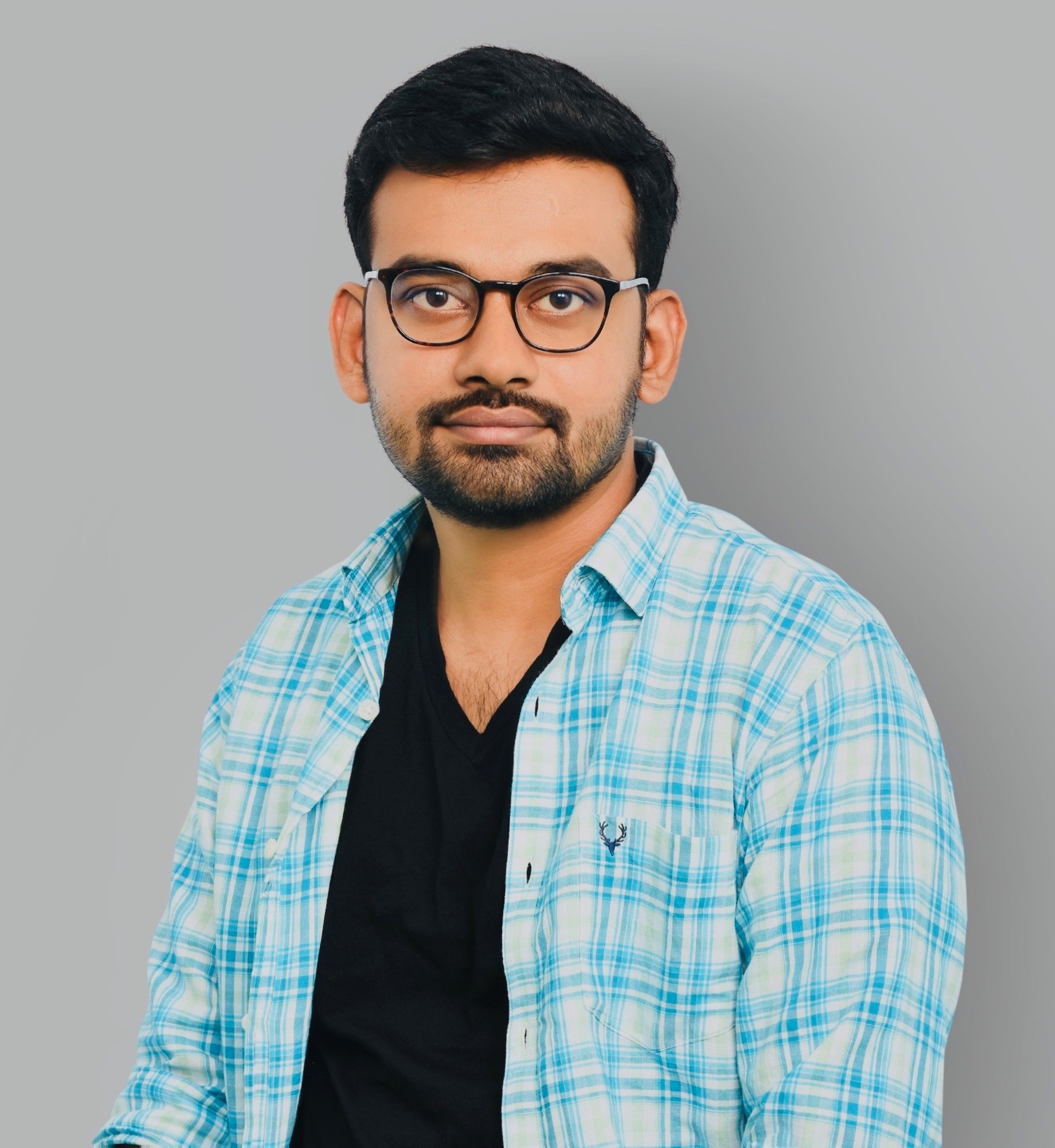}}]{Sudip Das} is a Senior Research Engineer in the department of Driving Software and Systems (DSW) at Valeo in India. He worked at Indian Statistical Institute in a research position in its Computer Vision and Pattern Recognition Unit at Kolkata after obtaining his B. Tech degree in Computer Science and Engineering from the West Bengal University of Technology (WBUT), India, in 2017. His principal area of research focuses on the relevant problems of Unsupervised Leaning, Curriculum Learning, Transfer Learning, and Domain Adaptation. He is also passionate to work on the various problems of autonomous driving. In particular, his research interests include Computer Vision and deep learning, with the goal of Detecting, Segmenting, and Pose Estimating of objects in Images or videos.

\end{IEEEbiography}

\begin{IEEEbiography}[{\includegraphics[width=1in,height=1.25in,clip,keepaspectratio]{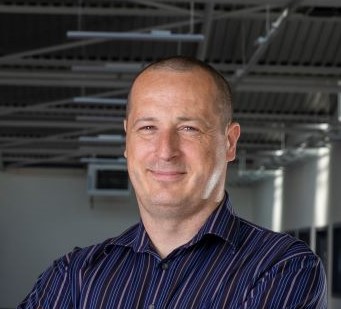}}]{Ciarán Eising}  received the B.E. in Electronic and Computer Engineering and the Ph.D. degree from the National University of Ireland in 2003 and 2010, respectively. From 2009 to 2020, he worked as a Computer Vision Team Lead and an Architect at Valeo Vision Systems, where he also held the title of Senior Expert. In 2016, he was awarded the position of Adjunct Lecturer at the National University of Ireland, Galway. In 2020, he joined the University of Limerick as a Lecturer in Artificial Intelligence and Computer Vision. Ciarán is a Senior Member of the IEEE.
\end{IEEEbiography}

\begin{IEEEbiography}[{\includegraphics[width=1in,height=1.25in,clip,keepaspectratio]{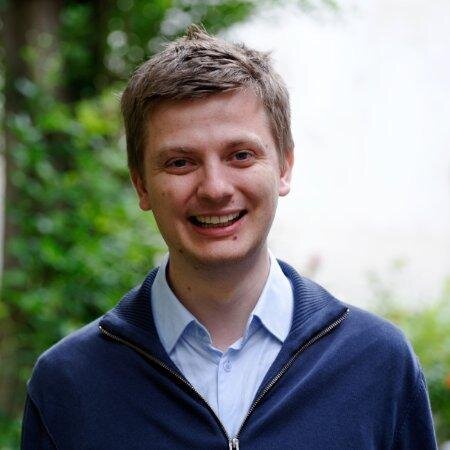}}]{Andrei Bursuc} is a research scientist at valeo.ai in Paris, France. He completed his Ph.D. at Mines ParisTech in 2012. He was a postdoc researcher at Inria Rennes and Inria Paris. In 2016, he moved to industry to pursue research on autonomous systems. His current research interests concern computer vision and deep learning, in particular annotation-efficient learning and predictive uncertainty quantification. Andrei serves regularly as a reviewer for major computer vision and machine learning conferences and journals. He is teaching undergraduate courses
at Ecole Normale Sup{\'e}rieure and Ecole Polytechnique.
\end{IEEEbiography}

\begin{IEEEbiography}
[{\includegraphics[width=1in,height=1.25in,clip,keepaspectratio]{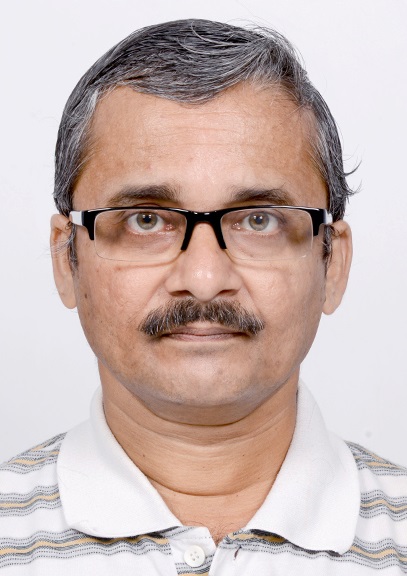}}]{Ujjwal Bhattacharya} is currently a member of the faculty of the Computer Vision and Pattern Recognition Unit of Indian Statistical Institute situated in Kolkata. He joined his current Institute in 1991 as a Junior Research Fellow after obtaining his M.Sc. and M. Phil. degrees in Pure Mathematics from Calcutta University. 
In the past, he collaborated with a few  industries and research labs in India and abroad. 
In 1995, he received Young Scientist Award from the Indian Science Congress Association. Also, he received a few best paper awards from various groups. He is a senior member of the IEEE and a life member of IUPRAI, the Indian unit of the IAPR. He has served as a Program Committee member of various reputed International Conferences and Workshops. 
Also, he worked as a Co-Guest Editor of a few Special Issues of International Journals. His current research interests include machine learning, computer vision, image processing, document processing, handwriting recognition, etc.
\end{IEEEbiography}

\begin{IEEEbiography}
[{\includegraphics[width=1in,height=1.25in,clip,keepaspectratio]{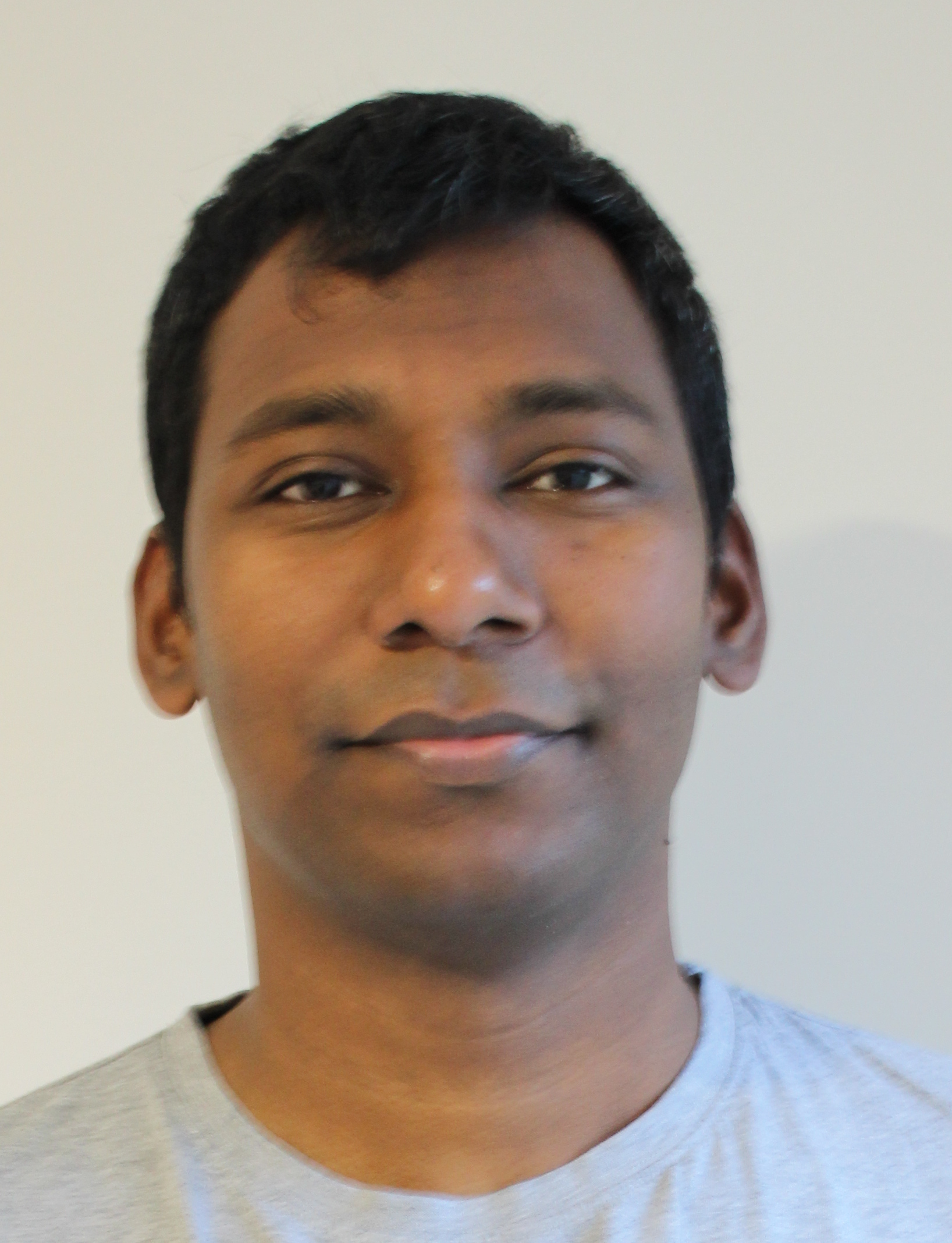}}]{Senthil Yogamani} is an Artificial Intelligence architect and holds a director-level technical leader position at Valeo Ireland. He leads the research and design of AI algorithms for various modules of autonomous driving systems. He has over 16 years of experience in computer vision and machine learning including 14 years of experience in industrial automotive systems. He is an author of 125+ publications with 5200 citations and 100+ filed patents. He serves on the editorial board of various leading IEEE automotive conferences including ITSC and IV and the advisory board of various industry consortia including Khronos, Cognitive Vehicles, and IS Auto. He is a recipient of the best associate editor award at ITSC 2015 and the best paper award at ITST 2012.
\end{IEEEbiography} 

\EOD

\end{document}